\documentclass[preprint,12pt]{elsarticle}

\usepackage{amssymb}
\usepackage{url}
\usepackage[ruled,lined,linesnumbered,longend]{algorithm2e}

\journal{Computers and Operations Research}

\def\codemaker{CM}
\def\codebreaker{CB}
\begin{document}

\begin{frontmatter}

\title{An experimental study of exhaustive solutions for the Mastermind puzzle}

\author[ugr,citic]{Juan-J. Merelo-Guerv\'os}
\author[ugr,citic]{Antonio M. Mora}
\author[uma]{Carlos Cotta}
\author[hi]{Thomas P. Runarsson}
\address[ugr]{Dept. of Computer Architecture and Technology,
  University of Granada, Spain, \\ email: \{jmerelo,amorag\}@geneura.ugr.es}
\address[citic]{CITIC, \url{http://citic.ugr.es}}
\address[uma]{ Dept. of Languages and Computer
Sciences, University of M\'alaga, \\ email: ccottap@lcc.uma.es}
\address[hi]{School of Engineering and Natural Sciences, University of Iceland,\\ email: tpr@hi.is }

\begin{abstract}
Mastermind is in essence a search problem in which a string of symbols that is kept secret must be
found by sequentially playing strings that use the same alphabet, and using the responses
that indicate how close are those other strings to the secret one as
hints. Although it is commercialized as a game, it is a combinatorial 
problem of high complexity, with applications on fields that range
from computer security to genomics. As such a kind of problem, there are no
exact solutions; even exhaustive search methods rely on heuristics to
choose, at every step, strings to get the best possible hint. These
methods mostly try to play the move that offers the best reduction in
search space size in the next step; this move is chosen according to an empirical score. However, in this paper we will examine several state of the art 
exhaustive search methods and show that another factor, the presence of the
actual solution among the candidate moves, or, in other words, the
fact that the actual solution has the highest score, plays also a very important
role. Using that, we will propose new exhaustive search approaches 
that obtain results which are comparable to the classic ones, and
besides, are better suited as a basis for non-exhaustive search
strategies such as evolutionary algorithms, since their behavior in a
series of key indicators is better than the classical algorithms. 

\end{abstract}


\end{frontmatter}

\section{Introduction and description of the game of Mastermind}
\label{s:intro}

Mastermind \cite{mm:mathworld} is a board game that has enjoyed
world-wide popularity in the last decades. Although its current version
follows a design created in the 70s by the Israeli engineer Mordecai Meirowitz
\cite{wiki:mm}, the antecedents of the game can be
traced back 
to traditional puzzles such as \emph{bulls and cows} \cite{francisstrategies:moo} or the so-called \emph{AB}
\cite{Chen2007435} game played in the Far East. Briefly, MasterMind is a two-player
code-breaking game, or in some sense a single-player puzzle, where one of the
players --the \emph{codemaker} (\codemaker)-- has no other role in the game than
setting a hidden combination, and automatically providing hints on
how close the other player --the \emph{codebreaker} (\codebreaker)-- has
come to correctly guess this combination. More precisely, the flow of the game
is as follows:
\begin{itemize}

\item The \codemaker\  sets and hides a length $\ell$ combination of $\kappa$
  symbols. Therefore, the \codebreaker\ is faced with $\kappa^\ell$
  candidates for the hidden combination, which is typically
  represented by an array of pegs of different colors (but can also be
  represented using any digits or letter strings) and hidden from the
  \codebreaker. 

\item The \codebreaker\  tries to guess this secret code by
  producing a combination (which we will call move) with the same length, and using the same set
  of symbols. As a response to that move, the \codemaker\, acting as an
      {\em oracle} (which explains the inclusion of this game in the
      category called {\em oracle} games)
  provides information on the number of symbols guessed in the right
  position (black pegs in the physical board game), and the number of
  symbols with the correct color, but in an incorrect position (white
  pegs); this is illustrated in Table \ref{tab:mm}.

\item The \codebreaker\  uses (or not, depending on the strategy he is
      following) this information to produce a new combination,
  that is assessed in the same way. If he correctly guesses
  the hidden combination in at most $N$ attempts, the \codebreaker\
      wins. Otherwise,  the \codemaker\ takes the game. $N$ usually
      corresponds to the 
      physical number of rows in the  game board, which is equal to
      fifteen in the first commercial version.

 \item \codemaker\ and \codebreaker\ are then interchanged, and several rounds of the game
       are played, this is a part of the game we do not
       consider. The player that is able to obtain the minimal amount
       of attempts wins. 

\end{itemize}

\begin{table}[t]
\caption{Progress in a MasterMind game that tries to guess the secret
  combination {\em ABBC}. 2nd and 4th combinations are not {\em
    consistent} with the  
  first one, not coinciding in two positions and one color
  with it. \label{tab:mm}}
  \centering\smallskip
\begin{tabular}{|l|c|}
\hline
\emph{Combination} & \emph{Response} \\
\hline
AABB & 2 black, 1 white\\
ABFE & 2 black\\
ABBD & 3 black\\
BBBE & 2 black \\
ABBC & 4 black\\
\hline
\end{tabular}
\end{table}
In this paper, we will consider the puzzle only from the point of view
of the \codebreaker\ , that is, a solver will be confronted with an
oracle that will award, for each move, a number of black and white
pegs. We will make no assumptions on the \codemaker\ other that it
will use random combinations (which in fact it does). A human player will probably have some
kind of bias, but we are interested in finding the most general
playing method
without using that kind of information. 

From the point of view mentioned in the last paragraph this puzzle is, in fact, a quite interesting combinatorial problem,
as it relates to other {\em oracle}-type problems such as the
 hacking of the PIN codes used in bank ATMs
 \cite{bank:mm,focardi2011guessing},  uniquely identifying a person
 from queries to a genetic database \cite{goodrich2009algorithmic}, or
 identifying which genotypes have interesting traits for selectively
 phenotyping them \cite{gagneur2011selective}; this is just a small
 sample of possible applications of the solution of the game of
 Mastermind. A more complete survey of applications (up to 2005) can
 be found in our previous paper \cite{mastermind05}, but recent posts
 in the questions-and-answers website StackOverflow are witness to the
 ongoing interest in finding solutions to the game
 \cite{stackoverflow}. 

Mastermind is 
also, as has been proved recently, a complex problem, paradigmatic of a whole class of search
problems \cite{o1991mastermind};  the problem of finding the solution has been shown to be
NP-complete under different formulations
\cite{abs-cs-0512049,Kendall200813}, and the problem of counting the
number of solutions compatible with a certain set of answers is
\#P-complete \cite{DBLP:journals/corr/abs-1111-6922}. 
This makes the
search for good heuristic algorithms to solve it a challenge; besides, several issues remain open,
such as what is the lowest average number of guesses needed to solve the
problem for any given $\kappa$ and $\ell$. Associated to this, there arises
the issue of coming up with an efficient mechanism for finding the
hidden combination independently of the problem size, or at least, a method that
scales gracefully when the problem size increases.

In this paper we will mainly examine empirical exhaustive search algorithms for
sizes in which they are feasible and try to find out which are the
main factors that contribute to finding the hidden combination in a
particular (and small) number of moves; in this context, we mean by
exhaustive algorithms those that examine, in turn, all items in the
search space, discarding a few in each step until the solution (the
secret code) is found. While most methods of
this kind consider as
the only important factor the reduction of the search space \cite{Kooi200513}, we will
prove that how different combinations, and in particular the secret
(and unknown) one is scored contributes also to finding the solution
faster.
Taking into account how often the hidden combination appears
among the top scorers for a particular method, we will propose new
exhaustive search methods that are competitive with the best
solutions for Mastermind known so far. Besides, these methods do not
add too much complexity to the solution and present certain behaviors
that are better than the empirical solutions used so far. 

It is obvious that exhaustive search methods can only be used for
small sizes in NP-hard problems such as this one. However, its
analysis and the proposal of new methods have traditionally been,
and will also be used, to design metaheuristic search methods (such as
evolutionary algorithms) that are able to search much further in the
parameter space. Since that has been our line of research for a long
time \cite{jj-ppsn96,genmm99,mastermind05}, eventually we will use them in those
algorithms, but that is outside the scope of this paper.

The rest of the paper is organized as follows: we lay out the
terminology and explain the solutions to the game of Mastermind in
Section \ref{s:mm}; the state of the art in solutions to it is
presented next, in Section \ref{s:soa}. Next we will analyze the
best performing Mastermind methods known so far in Section  \ref{s:sm}
and reach some conclusions on its way of working; the conclusions
extracted in this analysis will be used in Section \ref{s:nsm} to
propose and analyze the new solutions proposed in this paper, which we
have called Plus and Plus2. This Section will be followed by
conclusions in the last Section \ref{s:c}.

\section{The game of Mastermind}
\label{s:mm}

As mentioned in Section \ref{s:intro}, a MasterMind problem instance is
characterized by two parameters: the number of colors $\kappa$
and the number of pegs $\ell$. Let
$\mathbb{N}_\kappa=\{1,2,\cdots\,\kappa\}$ be the set of symbols
used to denote the colors. Subsequently, any combination, either the
hidden one or one played by the \codebreaker, is a string $c\in
\mathbb{N}_\kappa^\ell$. Whenever the \codebreaker\  plays a
combination $c_p$, a \emph{response} $h(c_p,c_h)\in\mathbb{N}^2$ is
obtained from the \codemaker, where $c_h$ is the hidden combination.
A response $\langle b, w \rangle$ indicates that the $c_p$ matches
$c_h$ in $b$ positions, and there exist other $w$ symbols in $c_p$
present in $c_h$ but in different positions.

Let us define {\em consistent} combinations \cite{o1991mastermind} in
the following way: a 
combination $c$ is \emph{consistent} with another played previously 
$c_p$ if, and only if, $h(c,c_p)=h(c_p,c_h)$, i.e., if $c$ has as
many black and white pegs with respect to the $c_p$ as $c_p$ has
with respect to the hidden combination. Intuitively, this captures
the fact that $c$ might be a potential candidate for secret code in light of the outcome of playing $c_p$. We can easily
extend this notion and denote a combination $c$ as consistent (or
feasible) if, and only if, it is consistent with all the combinations
played so far, i.e., $h(c,c^i_p) = h(c^i_p, c_h)$ for $1 \leqslant i \leqslant n$,
where $n$ is the number of combinations played so far, and $c^i_p$
is the $i-$th combination played. Any consistent combination is a
candidate solution. A {\em consistent set} is the set of all
consistent combinations at a particular point in the game. 

Consistent combinations are important because only a move using one of
them decreases the size of the consistent set, at least by one (the
combination itself). A non-consistent combination might or might not
do that, depending on its similarity to already played moves. So most
Mastermind strategies play a consistent combination. We will see next
how these concepts are used to find the secret combination in a
minimum number of combinations. 

\section{State of the art}
\label{s:soa}

As should be obvious, once the set (or a subset) of consistent
solutions has been found, different methods  have different heuristics
to choose which combination is played. A simple
strategy, valid at all levels, is to play the first consistent
combination that is found (be it in an enumerative search, or after
random draws from the search space, or searching for it using
evolutionary algorithms\cite{mastermind05}). In general, we will call
this strategy Random; in fact, the behavior of all these strategies is
statistically indistinguishable, and, even if it is valid, this strategy does not
offer the best results.

\begin{table*}[htb!]
\caption{Table of partitions after two combinations have been played;
  this table is the result of comparing each combination against all
  the rest of the set, which is the set of consistent combinations in
  a game after two combinations have already been played. In boldface,
  the combinations which have the minimal worst set size (which happen to be in the 1b-1w column, but it could be any one); in this
  case, equal to ten. A strategy that tries to minimize worst case
  would play one of those combinations. The column 0b-1w with all values equal
  to 0 has been suppressed; column for combination 3b-1w, being impossible, is
  not shown either. Some rows have also been eliminated for the sake
  of compactness.  \label{tab:partitions}}
\small
\centering
\tiny
\begin{tabular}{|l|c|c|c|c|c|c|c|c|c|c|}
\hline
Combination & \multicolumn{10}{|c|}{Number of combinations in the
  partition with response} \\
\hline
 &0b-2w& 0b-3w& 0b-4w& 1b-1w& 1b-2w& 1b-3w& 2b-0w& 2b-1w& 2b-2w& 3b-0w\\
AABA &0 & 0 & 0 & 14 & 8 & 0 & 13 & 1 & 0 & 3 \\
{\bf AACC} &8 & 0 & 0 & 10 & 5 & 0 & 8 & 4 & 1 & 3 \\
AACD &6 & 2 & 0 & 11 & 6 & 1 & 4 & 5 & 1 & 3 \\
AACE &6 & 2 & 0 & 11 & 6 & 1 & 4 & 5 & 1 & 3 \\
AACF &6 & 2 & 0 & 11 & 6 & 1 & 4 & 5 & 1 & 3 \\
{\bf ABAB} &8 & 0 & 0 & 10 & 5 & 0 & 8 & 4 & 1 & 3 \\
ABAD &6 & 2 & 0 & 11 & 6 & 1 & 4 & 5 & 1 & 3 \\
ABAE &6 & 2 & 0 & 11 & 6 & 1 & 4 & 5 & 1 & 3 \\
ABAF &6 & 2 & 0 & 11 & 6 & 1 & 4 & 5 & 1 & 3 \\
{\bf ABBC} &4 & 4 & 0 & 10 & 8 & 0 & 8 & 1 & 1 & 3 \\
ABDC &3 & 4 & 1 & 11 & 9 & 1 & 4 & 2 & 1 & 3 \\
ABEC &3 & 4 & 1 & 11 & 9 & 1 & 4 & 2 & 1 & 3 \\
ABFC &3 & 4 & 1 & 11 & 9 & 1 & 4 & 2 & 1 & 3 \\
ACAA &0 & 0 & 0 & 14 & 8 & 0 & 13 & 1 & 0 & 3 \\
{\bf ACCB} &4 & 4 & 0 & 10 & 8 & 0 & 8 & 1 & 1 & 3 \\
ACDA &0 & 0 & 0 & 16 & 10 & 2 & 5 & 3 & 0 & 3 \\
{\bf BBAA} &8 & 0 & 0 & 10 & 5 & 0 & 8 & 4 & 1 & 3 \\
{\bf BCCA} &4 & 4 & 0 & 10 & 8 & 0 & 8 & 1 & 1 & 3 \\
BDCA &3 & 4 & 1 & 11 & 9 & 1 & 4 & 2 & 1 & 3 \\
BECA &3 & 4 & 1 & 11 & 9 & 1 & 4 & 2 & 1 & 3 \\
BFCA &3 & 4 & 1 & 11 & 9 & 1 & 4 & 2 & 1 & 3 \\
{\bf CACA} &8 & 0 & 0 & 10 & 5 & 0 & 8 & 4 & 1 & 3 \\
{\bf CBBA} &4 & 4 & 0 & 10 & 8 & 0 & 8 & 1 & 1 & 3 \\
CBDA &3 & 4 & 1 & 11 & 9 & 1 & 4 & 2 & 1 & 3 \\
CBEA &3 & 4 & 1 & 11 & 9 & 1 & 4 & 2 & 1 & 3 \\
CBFA &3 & 4 & 1 & 11 & 9 & 1 & 4 & 2 & 1 & 3 \\
DACA &6 & 2 & 0 & 11 & 6 & 1 & 4 & 5 & 1 & 3 \\
EBAA &6 & 2 & 0 & 11 & 6 & 1 & 4 & 5 & 1 & 3 \\
FACA &6 & 2 & 0 & 11 & 6 & 1 & 4 & 5 & 1 & 3 \\
FBAA &6 & 2 & 0 & 11 & 6 & 1 & 4 & 5 & 1 & 3 \\
\hline
\end{tabular}
\end{table*}

And it does not do so because not all combinations in the consistent
are able to reduce its size in the next move in the same way. So other
not so naïve solutions concentrate on scoring all combinations in the
consistent set according to a heuristic method, and playing one of the
combinations that reaches the top score, a random one or the first one
in lexicographical order. This score is always based on the concept of Hash
Collision Groups, HCG \cite{Chen2007435} or {\em partitions} \cite{Kooi200513}.  All
combinations in the consistent set are compared with each other,
considering one the secret code and the other a candidate solution;
all combinations will be grouped in sets according to how they compare
to a particular one, as shown in Table \ref{tab:partitions}. For
instance, in such table there is a big  set of combinations whose
response is exactly the same: ABDC, ABED, ABFC, ABAD, ABAE,
ABAF... All these combinations constitute a {\em partition}, and their
score will be exactly the same. 

To formalize these ideas, let ${\vec \Xi}=\{\Xi_{ibw}\}$ be a
three-dimensional matrix that
estimates the number $\Xi_{ibw}$ of combinations that will remain feasible
after combination $c_i$ is played and response $\langle b, w\rangle$
is obtained from the \codemaker. Then, the potential strategies for the \codebreaker\ are:
\begin{enumerate}
    \item Minimizing the worst-case partition \cite{Knuth}: pick $c_i=\arg\min_i\{\max_{b,w}(\Xi_{ibw})\}$. For
instance, in the set in Table \ref{tab:partitions} this algorithm
would play one of the combinations shown in boldface (the first one in
lexicographical order, since it is a deterministic algorithm).
    \item Minimizing the average-case partition \cite{Berghman20091880,irving}: pick $c_i=\arg\min_i\{\sum_{b,w}p_{bw}\Xi_{ibw}\}$, where $p_{bw}$
    is the prior probability of obtaining a particular outcome. If for instance we compute $p_{bw} = \sum_i\Xi_{ibw}/\sum_{i,b,w}\Xi_{ibw}$, then AACD would be the combination played among those in Table \ref{tab:partitions}.
    \item Maximizing the number of potential partitions \cite{Kooi200513}: pick $c_i=\arg\max_i\{|\{\Xi_{ibw}>0\}|\}$, where $|C|$ is
the cardinality of set $C$.  For example, a combination such as ABDC in Table \ref{tab:partitions} would result in no empty partition. This strategy is also called \emph{Most Parts}.
\item Maximizing the information gained \cite{Neuwirth,bestavros,mm:ppsn:2010}: pick
    $c_i=\arg\max_i\{H_{b,w}\left(\Xi_{ibw}\right)\}$,
    where $H_{b,w}(\Xi_{i[\cdot][\cdot]})$ is the entropy of the corresponding
    sub-matrix. We will call this strategy Entropy. 
 \end{enumerate}

\begin{algorithm*}[htb!]
\caption{Choosing the next move in the general case}\label{alg:general}
\smallskip 

\textbf{typedef} Combination: \textbf{vector}$[1..\ell]$ \textbf{of} $\mathbb{N}_\kappa$\;
\BlankLine
\textbf{procedure} \textsc{NextMove} (\textbf{in:} $F$: List[Combination], \textbf{out:} $guess$: Combination)\;
\textbf{var} TopScorers: List[Combination]\; 
\textsc{Score}( $F$ )\;
$guess\leftarrow$ \textsc{RandomElement} ( TopScorers )\;
\end{algorithm*}
All strategies based on partitions work as follows:\begin{enumerate}
\item Score all combinations in the consistent set according to the
  method chosen (Entropy, Most Parts, Best Expected, or Minimize Worst)
\item Play one of the combinations with the best score
\item Get the response from the codemaker, and unless it is {\em all
    blacks}, go to the first step.
\end{enumerate}
This is also represented more formally in Algorithm \ref{alg:general},
where \textsc{Score} scores all combinations according to the
criterion chosen, and \textsc{RandomElement} extracts a random element
from the list passed as an argument using uniform distribution. 

Exhaustive strategies have only (as far as we know) been examined and compared for the
base case of $\kappa=6, \ell=4$, for instance in \cite{nicso};
restricted versions of the game have been examined for other spaces,
for instance in  \cite{o1991mastermind}. The best
results are obtained by Entropy and Most Parts, but its difference is
not statistically significant. All other results (including the
classical one proposed by Knuth \cite{Knuth}) are statistically
worse. That is why in this paper we will concentrate on these two
strategies, which represent the state of the art in exhaustive
search. Bear in mind that all these strategies are empirical, in the
sense that they are based in an assumption of how the reduction of the
search space work; there is, for the time being, no other way of
proposing new strategies for the game of mastermind. 

Besides, the size of the the search space is used as proxy for what is
actually the key to success of a strategy: the probability of drawing
the winning combination at each step. It is evident that reducing the
number of combinations will eventually result in drawing the secret
one with probability one. However, it should not be neglected that the
probability of drawing it even if it is not the only remaining
combination is non-zero and care should be taken so that this
probability is either maximized, or at least considered to minimize
the number of moves needed to find it. So far, and to the best of our
knowledge, no study has been made of this probability; we will show in
this paper its influence on the success of a strategy. 

Let us finally draw our attention to the first move. It is obviously
an important part of the game, and since, a priori, all combinations
are consistent a strategy would consist in using all combinations in
the game and playing one according to its score. However, this is not
a sensible strategy even for the smaller size; {\em consistency} does
not make any sense in absence of responses, since the partitions will
not hold any information on the secret code. So, in most cases, a
fixed move is used, and the one proposed by Knuth (according to its
own Minimize Worst strategy) is most usually employed. Since Knuth's
strategy \cite{Knuth} was created for $\kappa=6, \ell=4$ it is
difficult to extrapolate it to higher dimensions, so papers vary in
which combination is used. At any rate, the influence of the first
move will be felt mainly in the reduction of size achieved {\em
  before} the first empirical move is made (second move of the game),
but a good reduction will imply a significant change in the average
number of games. We will bear this in mind when testing the different
empirical strategies. 

Essentially, then, Most Parts and Entropy are the state of the art in
exhaustive search strategies. These algorithms, along with all others
used in this paper, are written in Perl, released under a open source
licence and can be downloaded from \url{http://goo.gl/G9mzZ}. All
parameters, experiment scripts, data extraction scripts and R data
files can be found within the {\tt app/IEEE-CIG} directory. Next we
will examine how these algorithms work in two different sizes, which
are considered enough to assess its performance; in
Section \ref{s:nsm} we will propose new methods and prove that they
obtain a better average number of moves making them the new state of
the art in exhaustive search algorithms for the game of Mastermind. 

\section{Analyzing exhaustive search methods}
\label{s:sm}

In this section, we will outline a methodology for analyzing
exhaustive solutions to the game of Mastermind and apply it to two
strategies that are usually considered the best, Entropy and Most
Parts, for two different problem sizes, keeping  $\ell=4$ fixed and
setting $\kappa=6$ (Subsection \ref{ss:k6}) and  $\kappa=8$
(Subsection \ref{ss:k8}).

\subsection{$\ell=4,\kappa=6$}
\label{ss:k6}

Let us look first at the smallest usual size, $\kappa=6$ and
$\ell=4$. The best two strategies in this case have been proved 
to be Entropy and Most Parts \cite{nicso,Berghman20091880}, in both cases
starting using Knuth's rule, ABCA \cite{Knuth}. An analysis of the number of moves obtained by each method
are shown in Table \ref{tab:me}.
\begin{table}[htb!]
\caption{Average (with error of the mean) and Maximum number of moves for two search strategies: Most Parts
  and Entropy. \label{tab:me}}
  \centering\smallskip
\begin{tabular}{|l|c|c|c|}
\hline
\emph{Method} & \multicolumn{3}{c|}{\emph{Number of moves}}\\
 & \emph{Average} & \emph{Maximum} & \emph{Median} \\
\hline
Entropy & 4.413 $\pm$ 0.006 & 6 & 4 \\
Most Parts & 4.406 $\pm$ 0.007 & 7 & 4 \\
\hline
\end{tabular}
\end{table}
To compute this average, 10 runs over the whole combination space were
made. In fact the difference in the number of moves is small enough to not
be statistically significant (using Wilcoxon test, p-value = 0.4577), 
but there is a
difference among them, the most striking being that, even if the
maximum number of moves is higher for Most Parts, Entropy has a higher
average. 

\begin{figure}[!htb]
\centering
\includegraphics{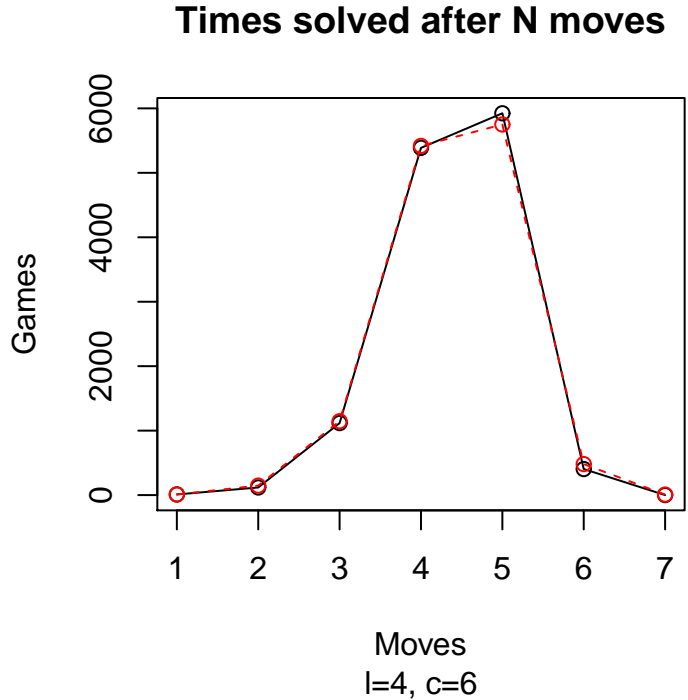}
\caption{Histogram with the count of the number of times every method
  is able to find the solution in every number of moves, for the
  Entropy method (black and solid) and Most Parts (light or red, dashed). \label{fig:histo:me}}
\end{figure}

To check where that difference lies we will plot a
histogram of the number of moves needed to find the solution for both
methods, see Fig. \ref{fig:histo:me}. 

%
\begin{figure}[!htb]
\centering
\includegraphics{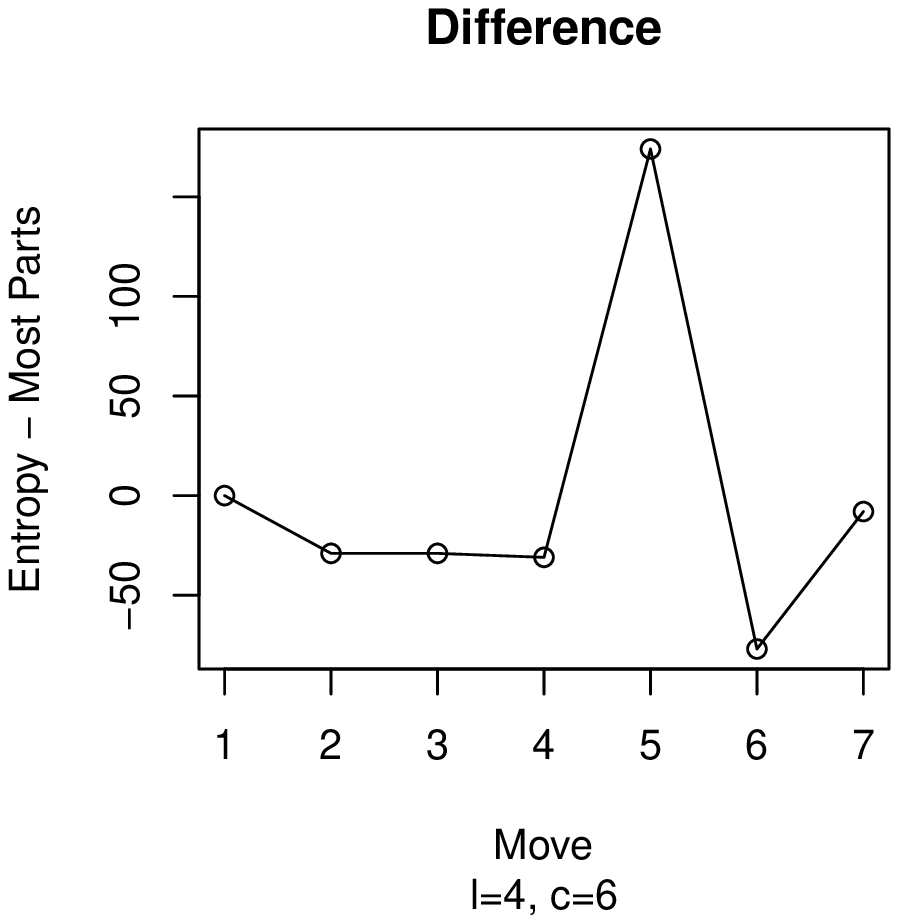}
\caption{. \label{fig:dif:me}}
\end{figure}
\begin{figure}[!htb]
\centering
\includegraphics{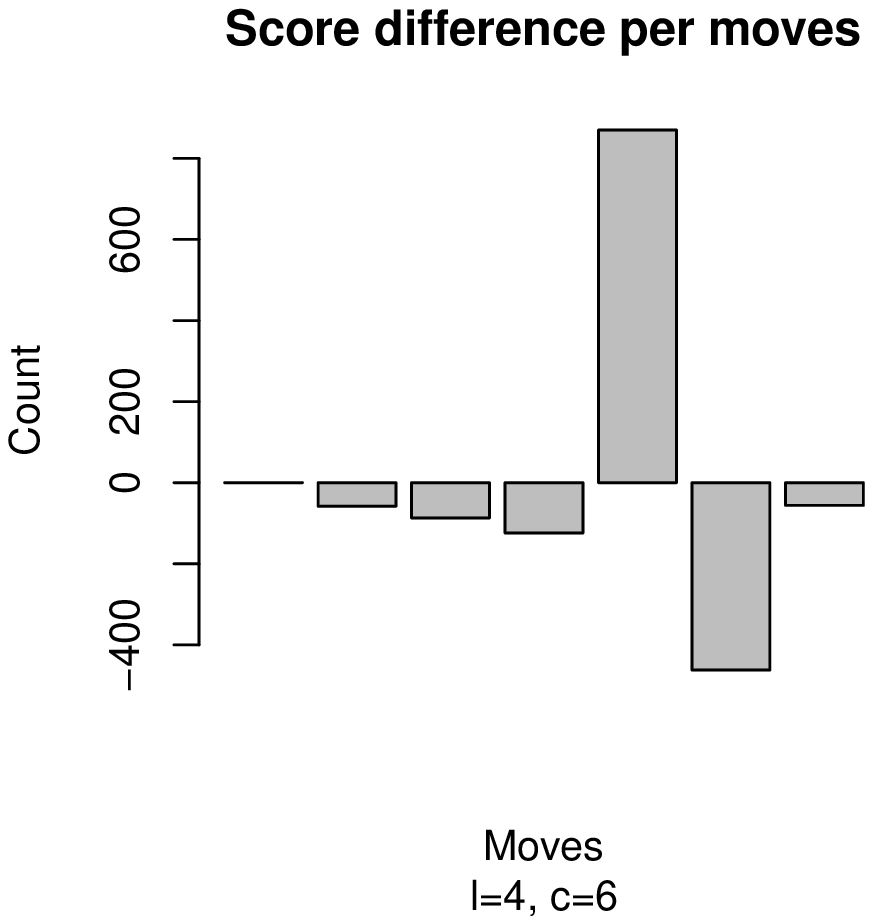}
\caption{Difference in score among the number of games won by Entropy and Most
  Parts in $x$ moves. Every $y$ value in the graph \ref{fig:dif:me}
  has been multiplied by the number of moves, resulting in the actual
  score the player would achieve. Remember that less is better;
  positive values mean that Entropy beats Most Parts (for that move),
  and vice versa. \label{fig:scoredif:me}}
\end{figure}

As should be expected for the negligible difference in the average number of moves, differences here
are very small. It is noticeable, however, than Most Parts finishes
less times in 1 to 3 moves, and also 5 moves. Most parts only finishes
in less occasions than Entropy for 4 moves; quite obviously, too,
there are a few times in which Most Parts needs 7 moves, but only
8 out of the total 12960 games (10 games times the total number of
combinations, 1296). Again, differences are not significant, 
but noticeable , so we will try to seek what is its source;
these differences must be the cause why eventually Most Parts achieves
a slightly higher average number of moves. Let us look further into
these differences by plotting the difference in the number of games
won by every one in a particular number of moves, let's say $x$. That
is shown in Figure \ref{fig:dif:me}. For all moves, except 5 and 7 (by
that move Entropy has always finished), Most Parts has more
games. Since both methods play the same number of games, the
difference is offset by the number of games more, around 150, that
Entropy finishes by move number five; this is also seen clearly in the
histogram \ref{fig:histo:me}. However, the important issue is the
difference in {\em score}, that is, in number of moves brought by
finishing each move. Differences in scores are computed by multiplying
the difference in the number of games by the move number, and plotted
in Figure \ref{fig:scoredif:me}. It obviously follows the same pattern
than Figure \ref{fig:dif:me}, however it gives us an idea of the
contribution of differences to total score. Since in this case the
total score for Entropy is 57189 and for Most Parts 57106, and Most
Parts is always better (negative difference in score) than Entropy, we
see that for low number of moves Most Parts is accumulating
differences, although it is worsened by not being able to finish as
many times as Entropy in 5 moves. This is the key move, and the shape
of the plot indicates a change of regime in the game. It also shows
that no single strategy is better than the other, yielding the
non-significant difference in moves (and scant difference in score,
less than 100 over 12960 games!). The main conclusion would be that ,
in fact, bot solutions are very similar, and this is also supported by
other experiments (not shown here) with more games, that do not yield
a significant difference either. However, we can also see that the way
they find the solution is different, that is why we will study other
aspects of the algorithm in the next paragraphs. 

\begin{figure}[!htb]
\centering
\includegraphics{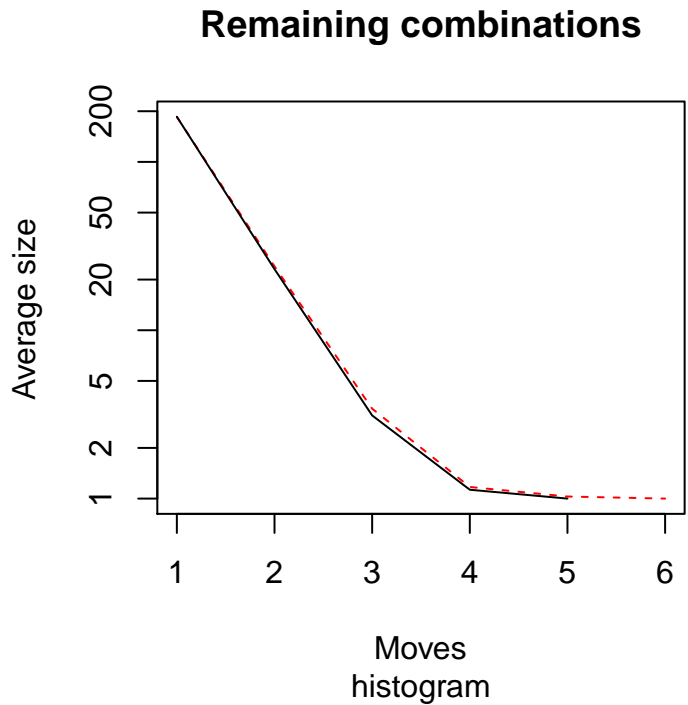}
\caption{Average number of combinations (or moves) remaining after
  each move, for the
  Entropy method (black and solid) and Most Parts (light or
  red). This plot corresponds to the numbers represented in Table
  \ref{tab:cset:me}. The $y$ axis is logarithmic for clarity, and the
  $x$ axis is shifted by one (1 means the second move, size before the
  first move is always the whole space).\label{fig:cset:me}}
\end{figure}

\begin{table}[h!tb]
\centering
\caption{Average and standard deviation of the number of combinations
  remaining after every move (or step). The numbers are the same after
  the  first move (not shown here), since they are playing the same one.\label{tab:cset:me}}
\smallskip
\begin{tabular}{|c|c|c|}
\hline
\emph{Before move \#} & \emph{Entropy} & \emph{Most Parts} \\
\hline
3 & 23 $\pm$ 14& 24 $\pm$  15\\
4 & 3.1 $\pm$ 1.8 & 3.4 $\pm$ 2.4 \\
5 & 1.13 $\pm$ 0.35 & 1.17 $\pm$ 0.43 \\ 
6 & 1 & 1.02 $\pm$ 0.17 \\
7 & & 1 \\
\hline
\end{tabular}
\end{table}
Since both methods try to reduce the size of the consistent set, we
will look at their size and how it changes with the number of
moves. We will log the size of the remaining consistent combinations
at every step, and this is shown in Fig. \ref{fig:cset:me} and Table
\ref{tab:cset:me}. 

In effect, the Entropy method is more efficient in
reducing the size of the set of remaining solutions, which is the
reason why usually it is presented as the best method for solving
Mastermind, since it achieves most effectively what it is intended to
achieve: reduction in search space size. At every step,
the difference is significant (using Wilcox test). This difference is
of a single combination at the beginning, and decreases with time while
still being significant; however, this reduction at the first stages
of the game makes the search simpler and would, in principle, imply an
easy victory for the Entropy technique, as would be expected. However, the
result is a statistical draw, 
with a slight advantage for the other
method, Most Parts. Besides, this reduction does not explain the
advantage found in Figures \ref{fig:histo:me} and \ref{fig:dif:me}: if the number of
solutions that remain is bigger (on average), why is Most Parts able
to find the solution at those same stages more often than Entropy,
which seems to be the key to success?

\begin{figure}[!htb]
\centering\smallskip
\includegraphics{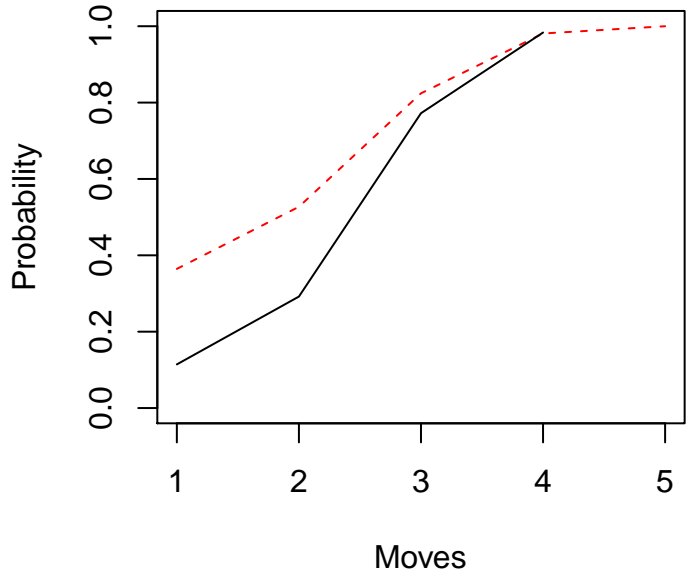}
\caption{Chance of finding the secret code among the top scorers for the
  Entropy method (black and solid) and Most Parts (light or
  red). This plot corresponds to the numbers represented in Table
  \ref{tab:top:me}.\label{fig:top:me}}
\end{figure}

To discover why this happen, we will look at another result. As explained above,
all methods are based on scoring consistent combinations according to
the partitioning of the set they yield, and then playing randomly one
of the combinations with the top score. If, by chance, the winning
combination is in that set of top scorers, there is a non-zero
possibility of playing it as next move and thus winning the match. In
effect, what we want is to maximize at every step the probability of
drawing the secret code. We
will then look at whether the winning combination effectively is or
not among the top scorers at each step. Results are shown in Fig. 
\ref{fig:top:me} and Table \ref{tab:top:me}.

\begin{table}[!htb]
\centering
\caption{Percentage of times the secret code is among the top scorers for
  each method.\label{tab:top:me}}
\smallskip
\begin{tabular}{|c|c|c|}
\hline
\emph{In move \#} & \emph{Entropy} & \emph{Most Parts} \\
\hline
2 & 0.1142857  & 0.3644788\\
3 & 0.2919495 & 0.5270987 \\
4 & 0.7721438 & 0.8242563 \\ 
5 & 0.9834515 &  0.9810066\\
6 & & 1 \\
\hline
\end{tabular}
\end{table}

Table \ref{tab:top:me} clearly shows that the probability of finding
the hidden combination among the top scorers increases with time, so
that in the 5th move is practically one. But it also shows that (in the first moves) there
is almost double the chance of having the winning combination among the top
ones for Most Parts than for Entropy, so, effectively, and obviously
depending on the size of the consistent set found at Table 
\ref{tab:cset:me}, the likelihood of playing the winning combination
is higher for Most Parts and its key to success. Making a back of the
envelope calculation, when the game arrives at the second move, which
roughly 11000 games do, a third of them will include the winning
combination among the top scorers, which again is roughly three
thousand. 

\begin{figure}[!htb]
\centering\smallskip
\includegraphics{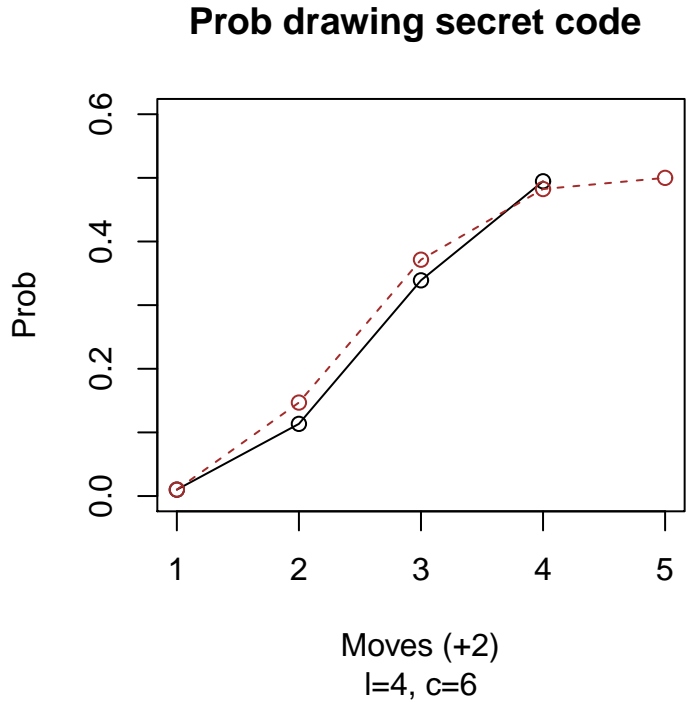}
\caption{Chance of drawing the secret code for the
  Entropy method (black and solid) and Most Parts (light or
  red). \label{fig:prob:me:46}}
\end{figure} 
This is not the whole picture, however. Every set of top scorers will
have a different size; even if the probability of finding the secret
score among that set is higher, the probability of {\em drawing} it
will be different if the size is smaller. The average of this
probability has been drawn in Figure \ref{fig:prob:me:46}. While
initially the probability is very small (but different), Most Parts
has a better chance of obtaining the secret code at each move than
Entropy, at least until move number 6 (label 4). That change of regime
is reflected in the previous plots by a sudden increase in the number
of games Entropy finishes (see Figures \ref{fig:histo:me} and
\ref{fig:dif:me}). However, since by that time Most Parts has been
able to finish a good amount of times, the difference is not too
big. Looking at figure \ref{fig:cset:me} and table \ref{tab:cset:me}, which plots the average size
of the set of remaining combinations, we see that by the 5th move
(label 4) there is, almost always, a single remaining combination, but
it happens more often for Entropy than for Most Parts; that is, in
this move is when the reduction of the search space effectively kicks
in, accounting for the {\em change in regime} we mentioned before. 

This, in turn, implies one of the main results of this paper: 
there are two factors in the success of a method for playing Mastermind. The
first is the reduction it achieves on the search space size by playing
combinations that reduce it maximally, but there is a second and
non-negligible factor: the chance of playing the winning
combination by having it among the top scorers. It can be said that
there is a particular number of moves, which in this case is after the
median, 4 moves, where a change of regime takes place. Before and up to 
that number of games, if a method finishes it is mainly due to
drawing, by chance, the secret code. After the median the secret code
is found because it is the only remaining element in the search
space. 

Most studies so far, however, had only looked at the smallest
size. Let us study another problem size, with search space 4 times as
big in the next subsection.

\subsection{$\ell=4,\kappa=8$}
\label{ss:k8}

The way the solutions work will change when the problem size is increased, so we
will perform the same measurements for problem size $\kappa=8, \ell=4$. Search space size is four times as big; time to solution grows faster
than lineally so it is not practical (although possible) to work with
exhaustive search for sizes bigger than that; in fact,
$\ell=5,\kappa=6$ is the next step that is feasible, but initial work
has shown that the behavior is not different from this case, and it
takes around two times as much; while the exhaustive
solution to the smallest Mastermind size considered takes around half
a second, it takes around 5 seconds for $\kappa=8, \ell=4$, and 10
seconds for $\kappa=6,\ell=5$. In practice, this means that instead of
using the whole search space 10 times over to compute this average, we will generate a
particular set of 5000 combinations, which includes all combinations
at least once and none more than two times. This instance set is available at
the method website (\url{http://goo.gl/ONYLF}).
A solution is searched, then, for every one of
these combinations. We are playing ABCD as first move, as one
interpretation of Knuth's \cite{Knuth} first move would say; that is,
we play half the alphabet and start again by the first letter (ABCA
would play half the ABCDEF alphabet and then start again). 

The average number of moves is represented in
Table \ref{tab:me:48}. This confirms in parts our above hypothesis:
the better capability Entropy has to decrease the size of the search
space gives it an advantage in the average number of moves, keeping at the
same time the ability of solving it in less maximum number of
moves. However, for this number of experiments, the difference is 
significant  (Wilcoxon paired test = 0.052). We will check
whether this difference has the same origin as we hypothesized
for the smaller search space size, by looking at the same variables. 

\begin{table}[htb]
\caption{Average (with error of the mean) and Maximum number of moves for two search strategies: Most Parts
  and Entropy for $\kappa=8, \ell=4$. \label{tab:me:48}}
  \centering
\smallskip
\begin{tabular}{|l|c|c|c|}
\hline
\emph{Method} & \multicolumn{3}{c|}{\emph{Number of moves}}\\
 & \emph{Average} & \emph{Maximum} & \emph{Median} \\
\hline
Entropy & 5.132 $\pm$ 0.012 & 8  & 5\\
Most Parts & 5.167 $\pm$ 0.012 & 8 & 5 \\
\hline
\end{tabular}
\end{table}
\begin{figure}[!htb]
\centering
\includegraphics{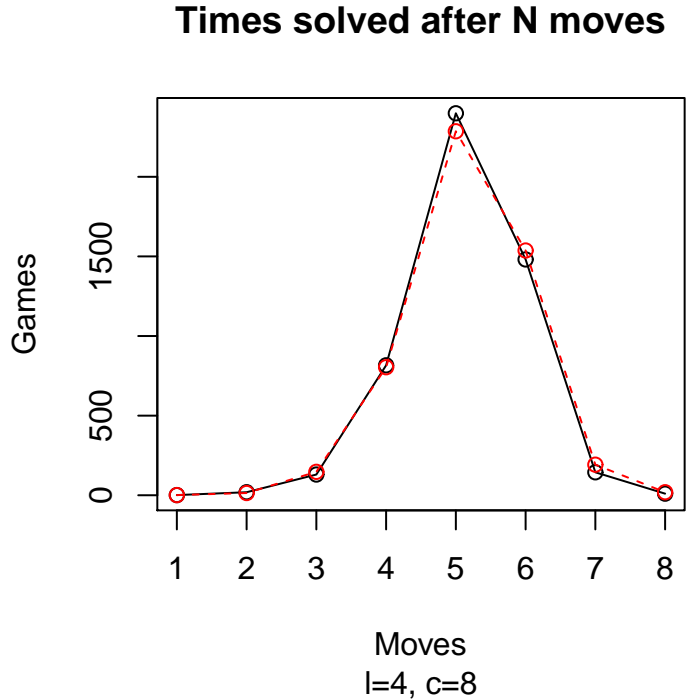}
\caption{Frequency of the number of moves up to (and including) the
  secret combination, for $\kappa=8, \ell=4$. As usual, red or light dashed line 
  represents Most Parts and solid black line Entropy. \label{fig:histo:me:48}}
\end{figure} 

\begin{figure}[!htb]
\centering
\includegraphics{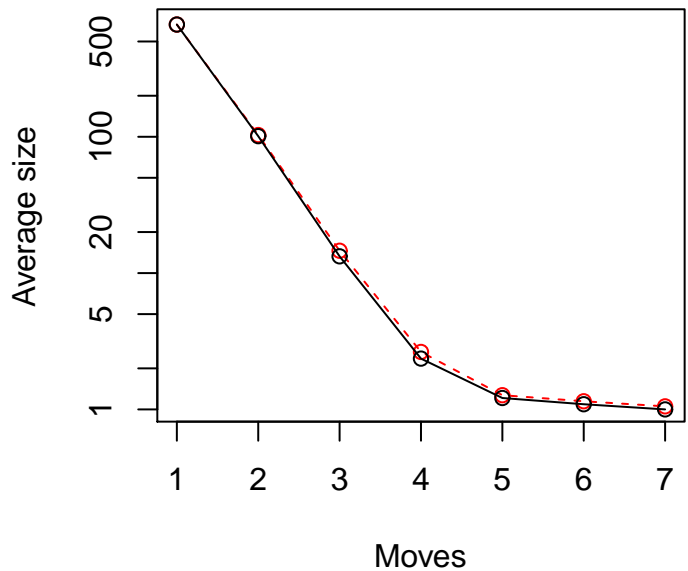}
\caption{Average size of the consistent set, that is, the set of
  solutions that have not been discarded at a point in the game for
  $\kappa=8, \ell=4$. As usual, red or light dashed line 
  represents Most Parts and solid black Entropy. \label{fig:cset:me:48}}
\end{figure}

The equivalent to Fig. \ref{fig:histo:me} has been represented in Fig \ref{fig:histo:me:48}.
The shape is similar, but the solid line that represents Entropy is slightly below the
dashed line for Most Parts for the highest number of moves which accounts for the small difference in
average number of moves. Is this difference accounted for by the size
of the search space after each move? As previously, we will plot it in
Fig. \ref{fig:cset:me:48} and Table \ref{fig:cset:me:48}. 
Differences for all moves are significant according to Wilcoxon test,
which means that actually Entropy is achieving what it was designed
for: reduce in a significant way the search space, until the secret
combination is found. However, in the same way as has been shown
above, other mechanisms are also at work to find the solution before
that reduction. 

\begin{table}[!htb]
\centering
\caption{Average and standard deviation of the number of combinations
  remaining after every move. The number of combinations is the same after
  the  first move (not shown here), since both are playing the same
  first move.\label{tab:cset:me:48}}
\smallskip
\begin{tabular}{|c|c|c|}
\hline
\emph{After move \#} & \emph{Entropy (ABCD)} & \emph{Entropy (ABCA)} \\
\hline
2 & 98 $\pm$ 67 & 102 $\pm$ 69 \\
3 & 13 $\pm$ 10 & 14 $\pm$ 12 \\ 
4 & 2.4 $\pm$ 1.6 & 2.63 $\pm$ 1.96 \\
5 & 1.21 $\pm$ 0.46 & 1.27 $\pm$ 0.52\\
6 & 1.07 $\pm$ 0.27 & 1.08 $\pm$ 0.31 \\
7 & 1 & 1.25 $\pm$ 0.46 \\
\hline
\end{tabular}
\end{table}
\begin{figure}[!htb]
\centering
\includegraphics{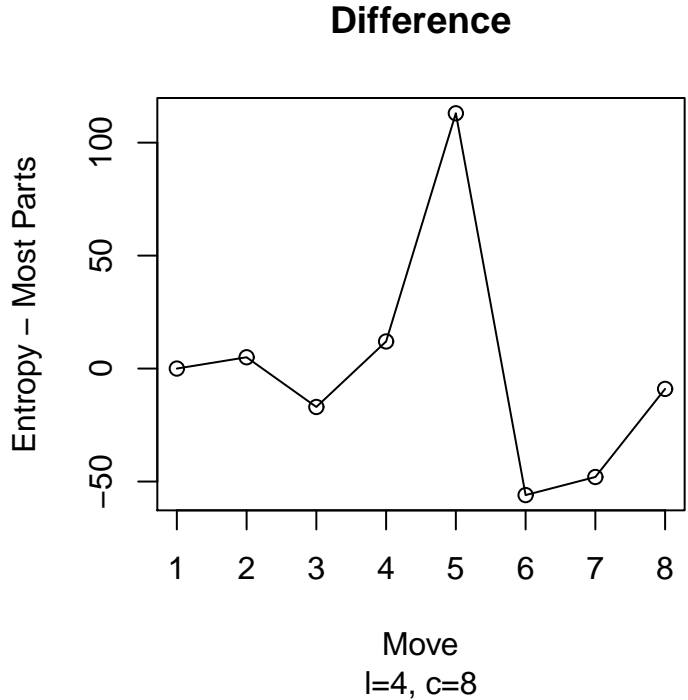}
\caption{Difference among the number of games won by Entropy and Most
  Parts in $x$ moves for  $\kappa=8, \ell=4$. \label{fig:dif:me:48}}
\end{figure} 
\begin{figure}[!htb]
\centering
\includegraphics{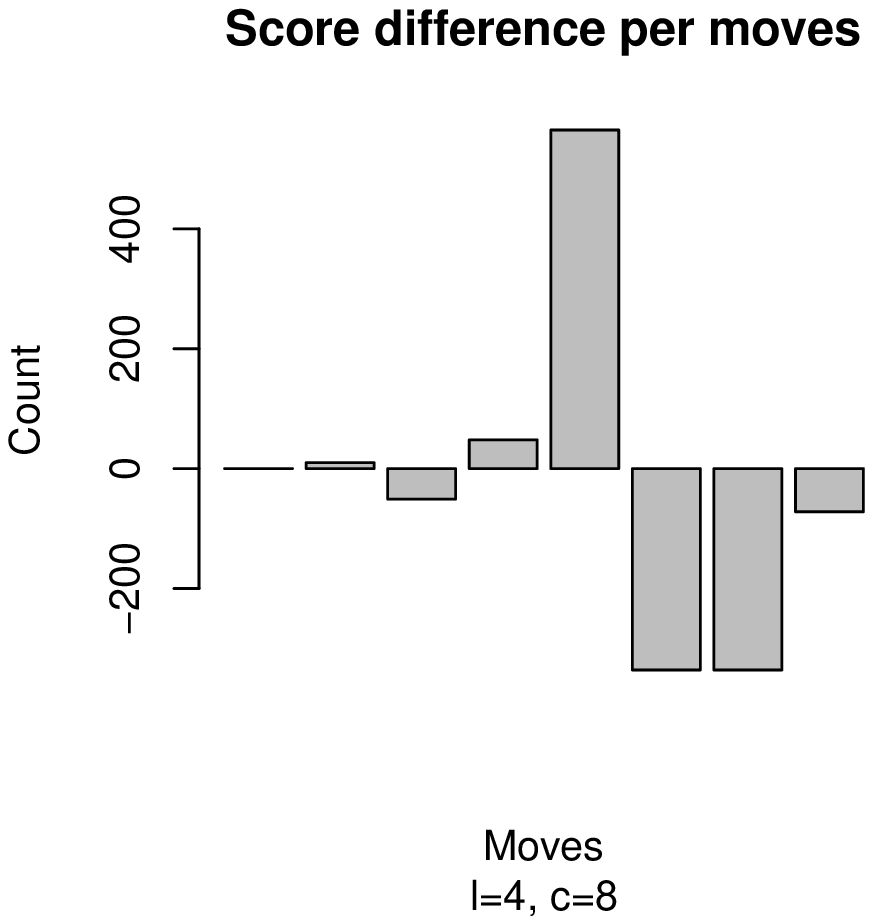}
\caption{Difference in score among the number of games won by Entropy and Most
  Parts in $x$ moves for  $\kappa=8, \ell=4$. Method is the same as
  for figure \ref{fig:scoredif:me}, that is, product of difference and
  number of moves. \label{fig:scoredif:me:48}}
\end{figure}

However, we will have to look in this case too at the differences in
games and score by both methods with is represented in Figures
\ref{fig:dif:me:48} and \ref{fig:scoredif:me:48}. The first one
represents the raw difference in number of times every one wins; again
there is a change in behavior or phase shift (which we have called
before a change of regime) when arriving at the fifth move, which is
also around the median (exactly the median, in this case; it was one
less than the median in the previous study). The scenario drawn in
\ref{fig:scoredif:me:48} is also similar, but shows that Entropy is
able to beat Most Parts mainly because it is able to finish more times
in less moves (around 5) than Most Parts (which accumulates lots of
bad games that need 6,7 and 8 moves). The final difference is around
200 points for the 5000 games (25834 vs. 25662) which is very small
but, in this case, significant. 

\begin{figure}[!htb]
\centering
\includegraphics{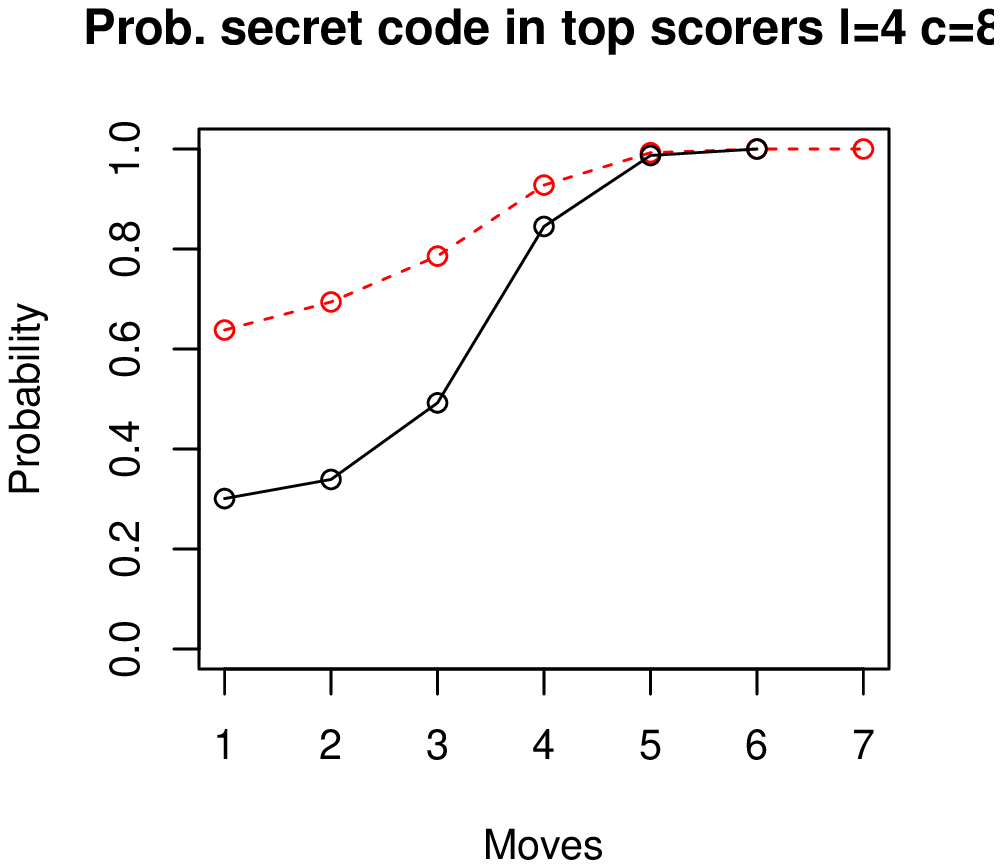}
\caption{Chance of finding the secret code among the top scorers for the
  Entropy method (black and solid) and Most Parts (light or
  red). This plot corresponds to the numbers represented in table
  \ref{tab:top:me:48} for $\kappa=8, \ell=4$.\label{fig:top:me:48}}
\end{figure}

We mentioned in Subsection \ref{ss:k6} that the probability of finding
the secret code among the top scorers was one of those features. It
should be expected that the probability of finding the secret code 
among the top scorers will change; since the number of elements in the
consistent set intuitively we should expect it to decrease. But this intuition is
wrong, as shown in Table \ref{tab:top:me:48} and Fig. \ref{fig:top:me:48}. 
In fact, if we compare these results with
those shown in Table \ref{tab:top:me} we see that, for the same move,
the proportion of times in which the secret code is among the top
scorers is almost twice as big at the beginning for Most Parts and
almost three times as big for Entropy. Interestingly enough, this also
implies that, while for $\kappa=6, \ell=4$ this probability was three
times as high for Most Parts, it is only two times as high now. 

\begin{table}
\centering
\caption{Percentage of times the secret code is among the top scorers for
  each method, $\kappa=8, \ell=4$.\label{tab:top:me:48}}
\begin{tabular}{|c|c|c|}
\hline
\emph{In move \#} & \emph{Entropy} & \emph{Most Parts} \\
\hline
2 & 0.3008602  & 0.6379276 \\
3 & 0.3425758 & 0.6957395 \\
4 & 0.4937768 & 0.7845089 \\ 
5 & 0.8630084 & 0.9160276\\
6 & 0.9898990 & 1 \\
\hline
\end{tabular}
\end{table}
%
\begin{figure}[!htb]
\centering\smallskip
\includegraphics{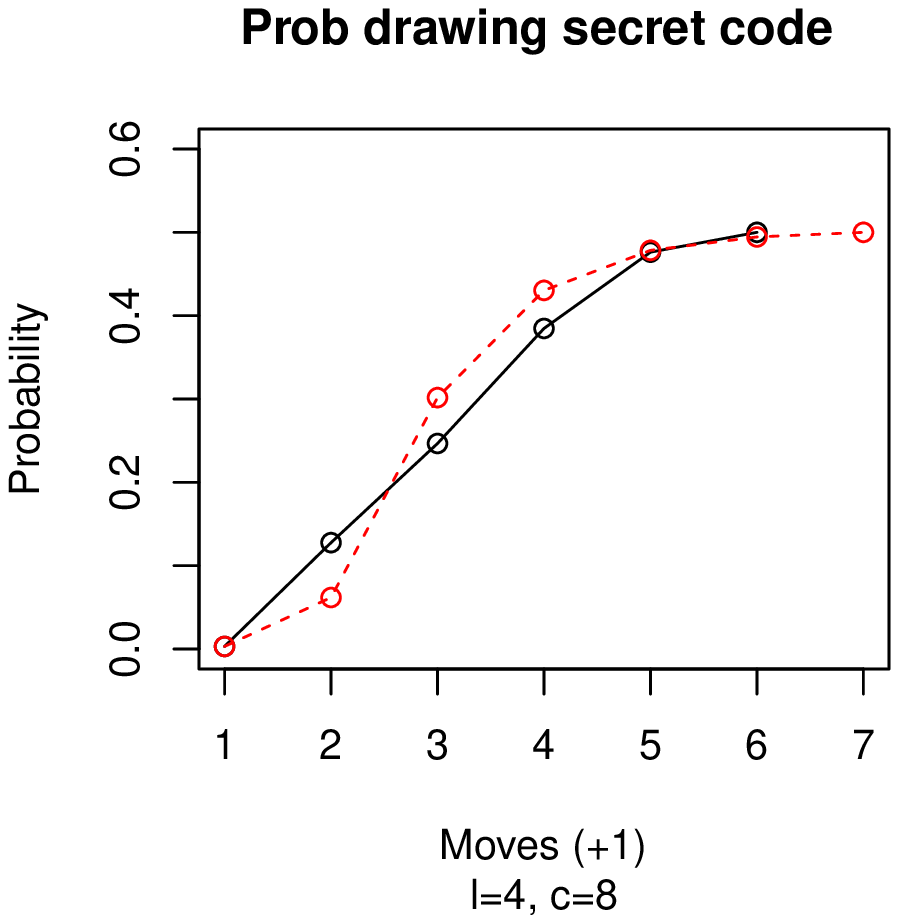}
\caption{Chance of drawing the secret code for the
  Entropy method (black and solid) and Most Parts (light or
  red) for $\kappa=8, \ell=4$. Remember that the $x$ axis is shifted
  by one, since the first move is fixed.\label{fig:prob:me:48}}
\end{figure} 

This decrease in the chance of finding the secret code might
explain the difference in the average number of moves needed to find
the solution, which for this size tilts the balance in the direction
of Entropy. While for the smaller size this probability was enough to
compensate the superior capability of the Entropy method in reducing
the size of the search space, in this case the difference is not so
high, which makes the Entropy method find the solution, on average, on
less moves. This is actually only one of the factors in the actual
probability of drawing the secret code, which is shown
in Figure \ref{fig:prob:me:48}. In the same fashion as the previous
study (shown in Figure \ref{fig:prob:me:46}), Most Parts is better
in playing the secret combination up to the sixth move (label 5 in the
graphic). But, after that, probabilities are similar and the change in
regime takes place; in fact, looking at \ref{fig:dif:me:48} and
\ref{fig:cset:me:48}, it could be argued that the fact that after the
6th move almost always there is only one combination left accounts for
the observed phase shift. 

As we concluded the previous section, it is non-negligible the
influence of the probability of a score that is able to consistently
give top score to the secret combination; however, in this case the
size of the search space implies that the probability of playing the
secret code is smaller (by the 3rd move in $\kappa=6$ it is as high as
the 4th move in $\kappa=8$), and thus its influence is not enough to
make Most Parts as good as Entropy.

At any rate, it is clear that, also for this size, the only
determinant factor is not the decrease in size brought by the method,
but also the ability to score correctly. This, in turn, can take us to
design new heuristic methods that are able to address both issues at
the same time, and also study the difference brought by a different
initial move. 


\section{New exhaustive search methods}
\label{s:nsm}

Our intention by proposing new methods is to use the two main factors
that we have proved have an influence in the performance of
Mastermind-playing methods to devise new strategies that could serve
as another empirical proof of the mechanism, and also, if possible,
obtain methods that are able to obtain better results (or, at least,
not worse results).

A new method for Mastermind will somehow have to match the two
capabilities featured by Entropy and Most Parts: the first reduces
competently the search space, and the second is able to score better
the actual secret combination, which can then be selected for playing
giving winning moves. So we will propose several methods that will try
to combine them with the objective of achieving a better average
number of moves.
\begin{algorithm*}[htb!]
\caption{Choosing the next move in the {\em Plus} Mastermind solution
  method. This algorithm is the new version we propose to the
  \textsc{NextMove} function presented in Algorithm \ref{alg:general}.}\label{alg:plus}
\smallskip
\textbf{typedef} Combination: \textbf{vector}$[1..\ell]$ \textbf{of} $\mathbb{N}_\kappa$\;
\BlankLine
\textbf{procedure} \textsc{NextPlus} (\textbf{in:} $F$: List[Combination], \textbf{out:} $guess$: Combination)\;
\textbf{var} TopScorersEntropy, 
 TopScorersMostParts, 
 TopScorersAll: List[Combination]\; 
\textsc{EntropyScore}( $F$ )\;
\textsc{MostPartsScore}( $F$ )\;
TopScorersEntropy$\leftarrow$ \textsc{TopScorersEntropy}( $F$ )\;
TopScorersMostParts$\leftarrow$ \textsc{TopScorersMostParts}( $F$ )\;
TopScorersAll $\leftarrow$ TopScorersEntropy $\cap$ TopScorersMostParts\;
$guess\leftarrow$ \textsc{RandomElement} ( TopScorersAll )\;
\end{algorithm*}

The first one, which we will call  {\em Plus}, works as follows (please see Algorithm \ref{alg:plus}): the set of
consistent combinations is scored according to {\em both} Most Parts
and Entropy. The sets of combinations with the top score according to
both methods are extracted. If its intersection is non-null, a random
combination from it is returned. If it is null, a random
string of the union of both sets is returned. 

What we want to achieve with this method is a reduction of the consistent set in the same way as Entropy, but, by
intersecting it with the set of top scorers for Most Part, the
probability of finding the winning combination among them is also
increased. 
\begin{algorithm*}[htb!]
\caption{Choosing the next move in the {\em Plus} Mastermind solution
  method. Please note that up to line 6, this algorithm is identical to \ref{alg:plus}.}\label{alg:plus2}
\smallskip
\textbf{typedef} Combination: \textbf{vector}$[1..\ell]$ \textbf{of} $\mathbb{N}_\kappa$\;
\BlankLine
\textbf{procedure} \textsc{NextPlus2} (\textbf{in:} $F$: List[Combination], \textbf{out:} $guess$: Combination)\;
\textbf{var} TopScorersEntropy, 
 TopScorersMostParts, 
 TopScorersAll: List[Combination]\; 
\textsc{EntropyScore}( $F$ )\;
\textsc{MostPartsScore}( $F$ )\;
TopScorersEntropy$\leftarrow$ \textsc{TopScorersEntropy}( $F$ )\;
TopScorersAll$\leftarrow$ \textsc{TopScorersMostParts}( TopScorersEntropy )\;
$guess\leftarrow$ \textsc{RandomElement} ( TopScorersAll )\;
\end{algorithm*}

There is no single way of combining the two scoring methods. A second
one tested, which we will call {\em Plus2} (and which is outlined in \ref{alg:plus2}), works similarly. It proceeds
initially as the Entropy method, by scoring combinations according to
its partition entropy. But then, the top scorers of this method are
again scored according to Most Parts. Out of the subset of top Entropy
scorers with the highest Most Parts score, a random combination is
returned. Please note that, in order to compute the Most Parts score
of the top Entropy scorers, the whole Consistent Set $F$ must be
scored. In \ref{alg:plus2}, line 7 would extract the top scorers
according to the Most Parts method from the set of top scorers by the
Entropy method. While in Plus what is played belongs
to a subset of Entropy and Most Parts only if their intersection is
non-zero, in Plus2 the new set of top scorers is {\em always} a subset
of the top scorers for Entropy, further scored using  Most Parts. 

As intended in Plus, Plus2 tries to reduce even more the consistent set size by
narrowing down the set of combinations to those that have only top
scores using both methods. In principle, the {\em priority} of Entropy and Most
Parts can be swapped, by first scoring according to Most Parts and
then choosing those with the best Entropy score. This method was also
tested, but initial results were not as good, so we will show only
results for these two and compare them with the traditional methods
analyzed in the previous Section \ref{s:sm}

\begin{figure}[!htb]
\centering
\includegraphics{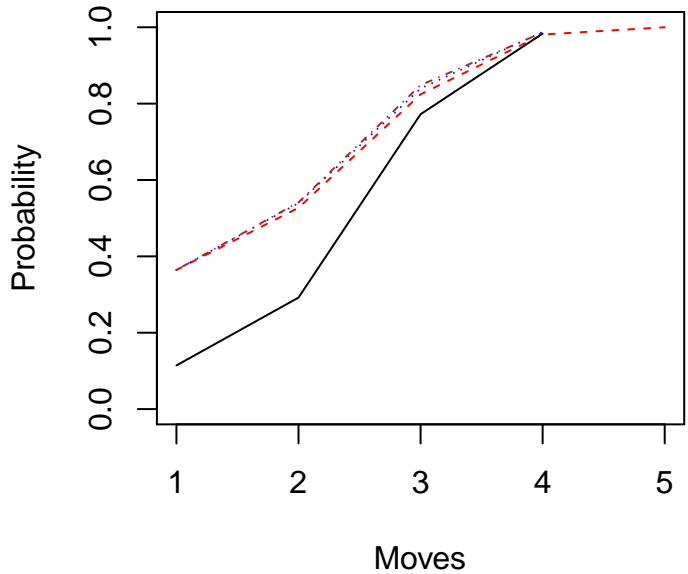}
\smallskip
\caption{Chance of finding the secret code among the top scorers for the
  Entropy method (black and solid) and Most Parts (light or
  red), to which we have added Plus (dotted, blue) and Plus2
  (dash-dotted, brown), for $\kappa=6, \ell=4$.\label{fig:top:plus}}
\end{figure}

Let us first look at the most important result: the number of
moves. They are shown in Table \ref{tab:plus:46}.
\begin{table}[htb]
\caption{Average and maximum number of moves for the new search
  strategies proposed, and comparison with the previous ones. \label{tab:plus:46}}
  \centering
\smallskip
\begin{tabular}{|l|c|c|}
\hline
\emph{Method} & \emph{Number of moves: Average} & \emph{Number of
  moves: Maximum} \\
\hline
Entropy & 4.413 $\pm$ 0.006 & 6 \\
Most Parts & 4.406 $\pm$ 0.007 & 7 \\
Plus & 4.404$\pm$ 0.007 & 6 \\
Plus2 & 4.41 $\pm$ 0.007 & 6 \\
\hline
\end{tabular}
\end{table}
Difference is not statistically significant, but as indicated in the
study presented in the previous section, it is encouraging to see that
results are, at least, as good as previously, and maybe marginally
better (for Plus, at least); in this particular Mastermind
competition, Plus would be the best; however, it is clear that
statistics dictate that it could happen otherwise in a different
one. At least the maximum number of moves is kept at the same level 
as Entropy, which might indicate that it achieves the reduction in
search space size we were looking for. In fact, the size of the
consistent set is practically the same than for Entropy. However,
there is some difference in the size of the sets of top scorers, which
is shown in Figure \ref{fig:top:plus} and Table \ref{tab:top:plus}.
\begin{table}
\centering
\caption{Percentage of times the secret code is among the top scorers for
  each method, $\kappa=4, \ell=6$. Entropy column has been suppressed
  for clarity. \label{tab:top:plus}}
\smallskip
\begin{tabular}{|c|c|c|c|}
\hline
\emph{In move \#} & \emph{Plus} &\emph{Plus2} & \emph{Most Parts} \\
\hline
2 &  0.364478 & 0.3644788 & 0.3644788\\
3 & 0.5395962&  0.5414171 & 0.5270987 \\
4 & 0.8397347& 0.8484542 & 0.8242563 \\ 
5 & 0.9877778 & 0.9862107 &  0.9810066\\
6 & & & 1 \\
\hline
\end{tabular}
\end{table}
\begin{figure}[!htb]
\centering\smallskip
\includegraphics{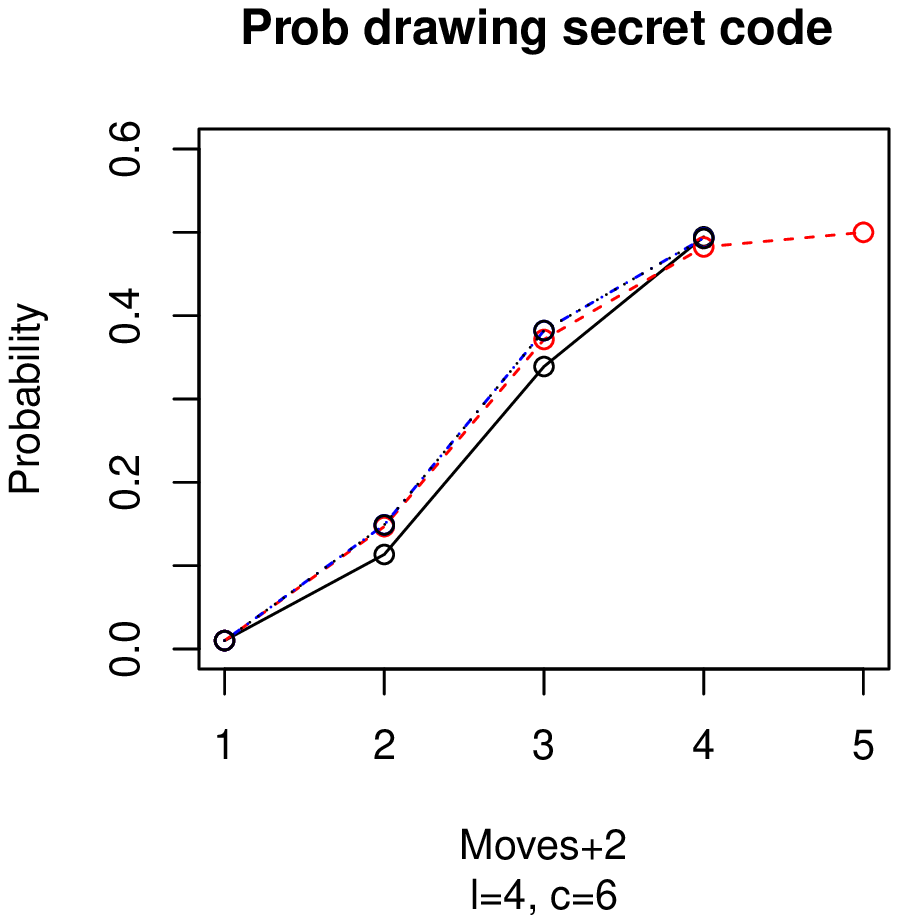}
\caption{Chance of drawing the secret code for the
  Entropy method (black and solid) and Most Parts (light or
  red) (same as in Figure \ref{fig:prob:me:46} to which we have added
  Plus and Plus2, practically one on top of the other and represented
  with dark-blue and dotted (Plus) and blue dash-dotted (Plus2). \label{fig:prob:plus:46}}
\end{figure} 
Table and graph show that, as intended, Plus and Plus2 reduce the size
of the consistent set  in the same proportion as Entropy does, but at the
same time, since the set of top scorers from which the move is
randomly chosen is smaller, the probability of finding the secret code
among them is higher; in fact, it is slightly higher than for Most
Parts. These two facts, together, explain the small edge in the number
of moves, which corresponds also to the small improvement in the
secret-playing probability shown in Figure \ref{fig:prob:plus:46}. In
fact, difference is significant from moves 1 to 3 (using Wilcoxon
test) for Plus2 against Most Parts, and from moves 2 and 3 for
Plus. Difference in total score is very small between the best (Plus,
57072) and the worst (Entropy, 57189), so, essentially, all methods
could obtain the same results. However, we have achieved
 to (significantly) increase the probability of obtaining the secret
 combination at each step, which might account for the small
 difference. Since this difference is offset by other random factors,
 however, no method is significantly better than other, and the
 p-value comparing Plus and Entropy is only 0.2780.

As we have seen before, the scenario is different at other sizes, so we will have to experiment them over the $\kappa=8,
\ell=6$ space. We use the same instance of 5000 combinations as we did
in the previous Subsection \ref{ss:k8}. This size was big enough to find some differences among
methods, and small enough for being able to perform the whole
experiment in a reasonable amount of time (around 90 minutes for the
whole set). The average number of moves is shown in Table
\ref{tab:plus:48}.
\begin{table}[htb]
\caption{Average and maximum number of moves for the two new search
  strategies, Plus and Plus2, along with the previously shown Most Parts
  and Entropy for $\kappa=8, \ell=4$. \label{tab:plus:48}}
  \centering
\smallskip
\begin{tabular}{|l|c|c|c|}
\hline
\emph{Method} & \multicolumn{3}{c|}{\emph{Number of moves}}\\
 & \emph{Average} & \emph{Maximum} & \emph{Median} \\
\emph{Method} & \emph{Number of moves: Average} & \emph{Number of
  moves: Maximum} \\
\hline
Entropy & 5.132 $\pm$ 0.012 & 8  & 5\\
Most Parts & 5.167 $\pm$ 0.012 & 8 & 5 \\
Plus & 5.154 $\pm$ 0.012 & 8 & 5\\
Plus2 & 5.139  $\pm$ 0.012 & 8 & 5\\
\hline
\end{tabular}
\end{table}
While previously the difference between Most Parts and Entropy was 
statistically significant, the difference now between Most Parts and
Plus2 is significant with p= 0.08536. It is not significant the
difference between Plus/Plus2 and
Entropy, Plus and Most Parts, and obviously between Plus and Plus2. This is, indeed, an
interesting result that shows that we have been able to design a
method that statistically is able to beat at least one of the best
classical methods, Most Parts, however, the edge obtained by them is
not enough to gain a clear victory over both of them. 
\begin{figure}[!htb]
\centering
\includegraphics{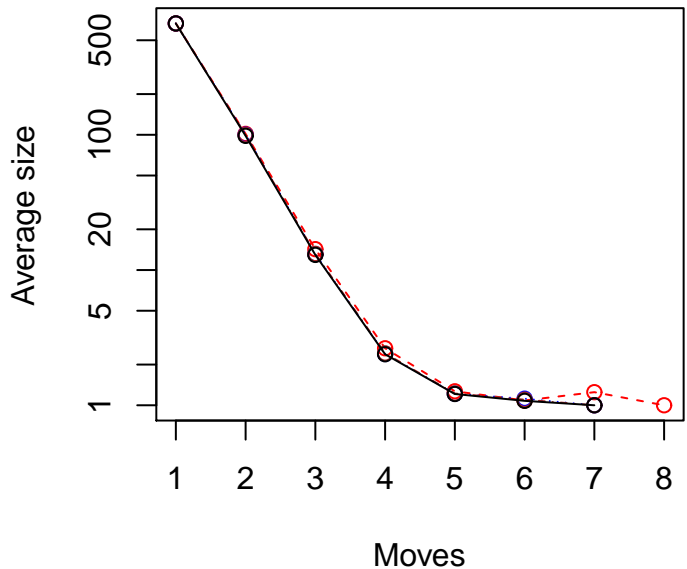}
\caption{Average size of the consistent set, that is, the set of
  solutions that have not been discarded at a point in the game for
  $\kappa=8, \ell=4$. As usual, red or light dashed line 
  represents Most Parts and solid black Entropy, dotted and blue for
  Plus and dash-dotted and brown for Plus2. Please note that the $y$
  axis is logarithmic; we have put it that way to highlight
  differences. \label{fig:cset:plus:48}}
\end{figure} 
\begin{figure}[!htb]
\centering
\includegraphics{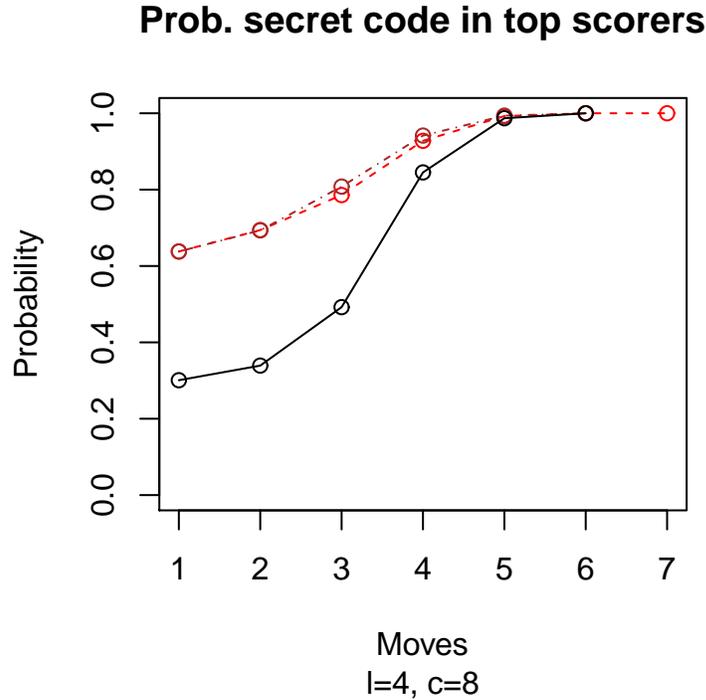}
\caption{Chance of finding the secret code among the top scorers for
  the Plus (dotted, blue), Plus2 (dash-dotted, brown),
  Entropy method (black and solid) and Most Parts (light or
  red).\label{fig:top:plus:48}}
\end{figure} 
\begin{figure}[!htb]
\centering
\includegraphics{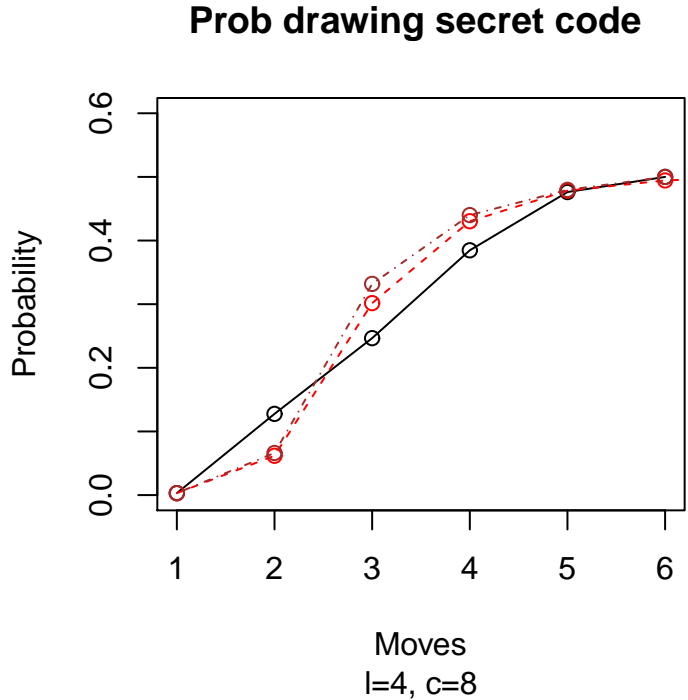}
\caption{Chance of finding the secret code among the top scorers for the
  Entropy method (black and solid) and Plus2 (light or
  brown, dot-dashed). Plus has been suppressed for clarity; the $x$
  axis is shifted by one, with $x=1$ representing the second
  move.\label{fig:prob:plus:48}} 
\end{figure} 

Let us check whether this difference stems from the design of the new
algorithms by looking, as before, at the size of the consistent sets
(Fig. \ref{fig:cset:plus:48} 
and the probability of finding the secret code among the top scorers
(Fig. \ref{fig:top:plus:48}. Once again, the difference among the new
methods and the old ones is very small, and almost none between
them. The average set size is virtually the same as Entropy (98 vs. 99
and 100 after the second move, for instance), but this was to be
expected; they are, anyway, smaller than the size of the sets for Most
Parts. 

The {\em finding the secret code by chance} probability is, as
intended, more similar to Most Parts (dotted and dot-dashed lines over the
red dashed line); that is, the probability of finding the secret code
among the top scorers is higher than for Entropy, also as
intended. This mixed behavior (reduction of search space as in
Entropy, probability of finding the secret code among the top scorers
as in Most Parts) explain why the methods proposed in this paper reach
the results shown in Table \ref{tab:plus:48}, which improve, in a
significant way, the state of the art in solutions of the game of
MasterMind. But looking more precisely at the actual chance of drawing
the secret code (not the binary probability of finding it among the
top scorers as in the previous graph), represented in Figure
\ref{fig:prob:plus:48} we find the situation very similar to the one
shown in \ref{fig:prob:me:48}, however, the probability of drawing the
secret code is better for Plus2 than for Most Parts, and significantly
so for all moves up to the 5th. However, it is still worse for Plus2
than for Entropy for the third move ($x=2$), which accounts for its
eventual tie with it. However, the difference with respect to Most
Parts seems to be enough to achieve a small difference at the end. 

As a conclusion to this section, we have proposed two new methods for
solving mastermind that try to improve the capability of drawing the
secret combination at the beginning of the game while, at the same
time, decreasing the size of the search space as Entropy does. The
proposed methods obtain results that are better than the worst of the
previous methods (Entropy or Most Parts), but not significantly better
than the best method (Most Parts or Entropy). However, they are robust
in the sense than they are at least as good (statistically) as the
best for each size, so in a sense it could be said that we have
achieved a certain degree of success. Let us see if this conclusion
holds by slightly changing the circumstances of the game by using a
different initial combination. 

\section{Studying the effect of a different starting combination on the differences among algorithms}

To test whether, under different circumstances, the newly proposed
methods perform as well as the best one, and also check the influence
of changing the initial combination to a different interpretation of
Knuth's first move, we will compare the two best algorithms seen
before, Plus2 and Entropy for $\kappa=8$ using ABCA as first move (instead of the
previously used ABCD). In principle, if we apply the partition score
to the first move (which can be done, even in the absence of a
reduction of search space brought by a move) a combination with
different symbols (such as ABCD) obtains the maximum entropy
encore. However, in the absence of information about the secret
combination it is again a empirical exercise to test different initial
moves, as has been done in \cite{Berghman20091880}, for instance. 

\begin{table}[htb]
\caption{Comparison of average and maximum number of moves using ABCA as first move and the best of the previous ones. \label{tab:abca:48}}
  \centering
\smallskip
\begin{tabular}{|l|c|c|c|}
\hline
\emph{Method} & \multicolumn{3}{c|}{\emph{Number of moves}}\\
 & \emph{Average} & \emph{Maximum} & \emph{Median} \\
\hline
Most Parts & 5.167 $\pm$ 0.012 & 8 & 5 \\
Entropy (ABCD) & 5.132 $\pm$ 0.012 & 8  & 5\\
Plus2 (ABCD) & 5.139  $\pm$ 0.012 & 8 & 5\\
Entropy (ABCA) & 5.124 $\pm$ 0.012 & 8  & 5\\
Plus2 (ABCA) & 5.116  $\pm$ 0.012 & 8 & 5\\
\hline
\end{tabular}
\end{table}
The summary of the number of moves is again shown in Table
\ref{tab:abca:48}. {\em A priori}, the average number of moves is
better than before. However, the only significant difference (as
usual, using paired Wilcoxon test) is
between Entropy(ABCA) and Plus2(ABCA) and Most Parts and Plus. The
difference between the ABCA and ABCD versions of both algorithms is not
significant. The difference between Plus2 and Entropy using ABCA as
first move is, once again, not significant, proving that the new
algorithm proposed, plus2, is as good as the best previous algorithm
available for the size (which, in this case, is Entropy). 

We can also conclude from this experiment that, even if there is not a
significant difference for a particular algorithm in using ABCD or
ABCA, it is true that results using ABCA are significantly better for
Entropy and Plus2 than for other algorithms such as Most Parts, with
which there was no significant difference using ABCD. We have not
tested Most Parts and Plus in this section since our objective was
mainly comparing the best methods with a new starting move with all
methods using the other move; however, we can more or less safely
assume that results will be slightly, but not significantly, better,
and that they will be statistically similar to those obtained by
Entropy and Plus2. 
\begin{figure}[!htb]
\centering
\includegraphics{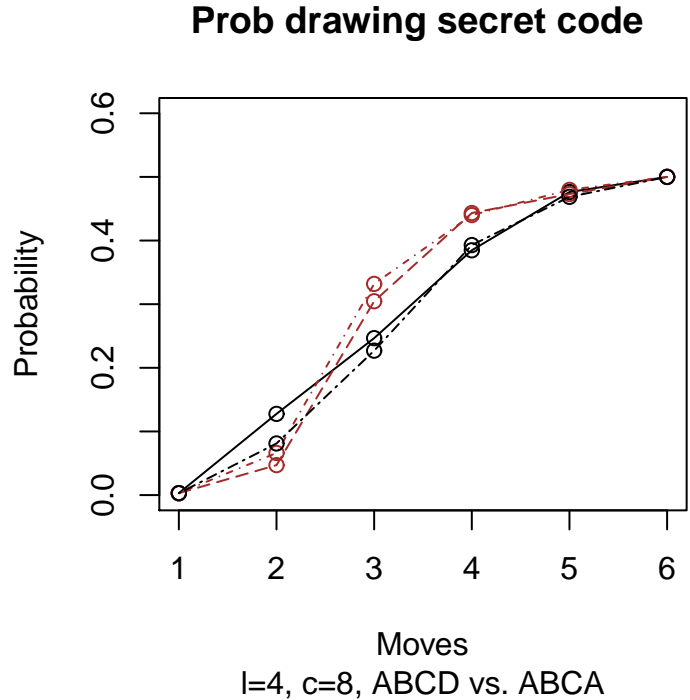}
\caption{Chance of finding the secret code among the top scorers for the
  Entropy method (black and solid) and Plus2 (light or
  brown, dot-dashed), both for ABCD as first move, and Entropy (black,
  dot-long-dash) and Plus2 (brown or light, long dash) for ABCA .\label{fig:prob:abca:48}} 
\end{figure} 

\begin{table}[!htb]
\centering
\caption{Average and standard deviation of the number of combinations
  remaining after every move for Entropy ABCD and ABCA (the quantities
  for Plus2 are practically the same). The number of combinations is the same after
  the  first move (not shown here), since both are playing the same
  first move.\label{tab:cset:abca:48}}
\smallskip
\begin{tabular}{|l|c|c|}
\hline
\emph{Before move \#} & \emph{Entropy ABCD} & \emph{Entropy ABCA} \\
\hline
2 & 666 $\pm$ 310 & 706 $\pm$ 327 \\
3 & 101 $\pm$ 67 & 99 $\pm$ 69 \\
4 & 13 $\pm$ 10 & 13 $\pm$ 10 \\ 
5 & 2.4 $\pm$ 1.5 & 2.28 $\pm$ 1.48 \\
6 & 1.21 $\pm$ 0.46 & 1.23 $\pm$ 0.5\\
7 & 1.09 $\pm$ 0.29 & 1.19 $\pm$ 0.39 \\
\hline
\end{tabular}
\end{table}
\begin{figure}[!htb]
\centering
\includegraphics{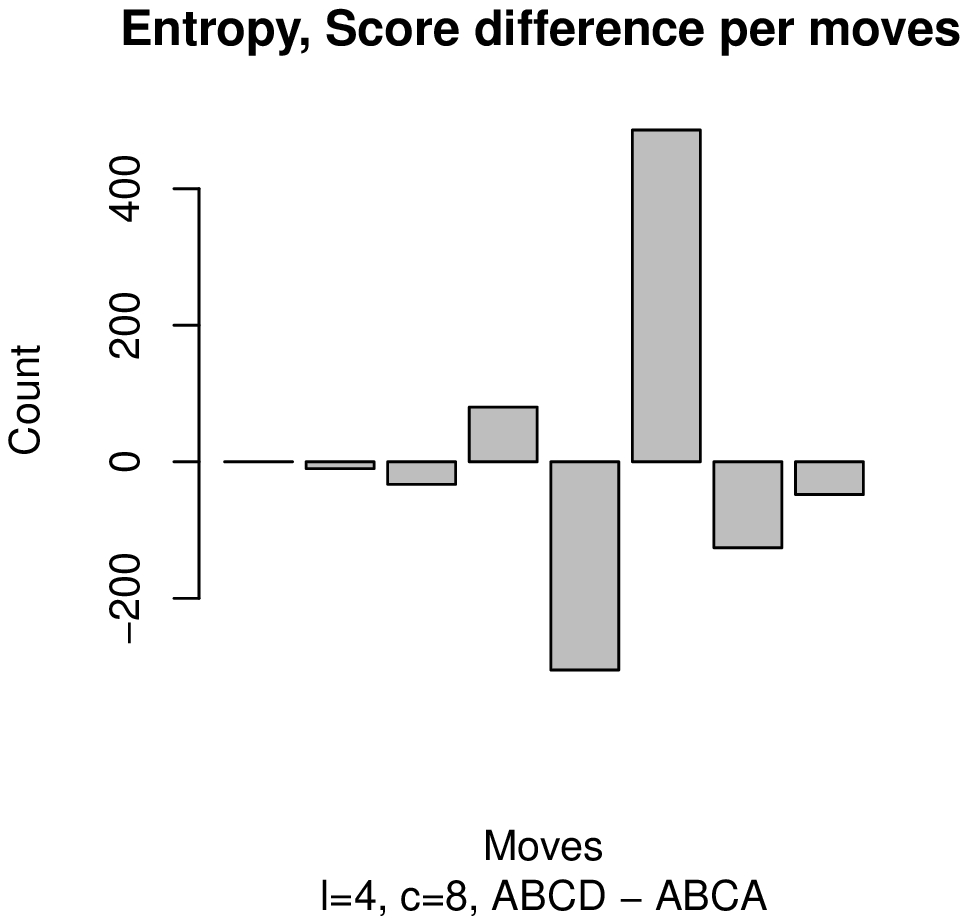}
\caption{Difference in score among the number of games won by Entropy
  (ABCD) and Entropy (ABCA)  in $x$ moves for  $\kappa=8, \ell=4$. Method is the same as
  for figure \ref{fig:scoredif:me}, that is, product of difference and
  number of moves. \label{fig:scoredif:abca:48}}
\end{figure}

It is clear that, in this case, the advantage is mainly due to the
changes in search space from the first move, however, this will have
an influence on the probability of playing the secret combination, as
shown in Figure \ref{fig:prob:abca:48}. In both cases (Entropy and
Plus2) the probability is {\em lower}, however, the result is (not
significantly) better. The differences must be due, then, mainly to
the difference in consistent set size, which is shown in Table
\ref{tab:cset:abca:48}, but these differences are not so clear-cut as
should be expected, that is, smaller sizes throughout all moves. The
size is actually bigger for ABCA in the first move, although smaller
in the second. These two will actually have little influence on the
outcome (since, at that stage in games, drawing the secret code is
mainly the product of the scoring algorithm and the {\em composition}
of the top scorers). However, there is a significant difference before
move 5th, which is when in some cases the consistent set is reduced to
one, which is also reflected in the differences in score represented
in Figure \ref{fig:scoredif:abca:48}. Up to the fifth move (label 2 in
the probability graph \ref{fig:prob:abca:48}, Entropy(ABCD) is better
(negative difference, Entropy(ABCD) $<$ Entropy(ABCA). However, it
accumulates score (remember that higher score is worse) in the 6th
move, which eventually implies the victory (which we should note is
not significant) of Entropy (ABCA). 

At any rate, in this section we have proved that the two mechanisms we
studied in the previous sections is also at work to provoke the
advantage of one algorithm over other, and that studying mainly
consistent set size and the probability of drawing the secret code,
and when one mechanism for finishing the game takes over the other, is
the best way of evaluating different algorithms for playing the game
of mastermind.

\section{Conclusions and discussion}
\label{s:c}

In this work we have analyzed the solutions to the game of Mastermind
in a novel way, taking into account not only the reduction of the size
of the search space brought by the combination played, but also how
every method scores the components of the search space and whether the
secret combination is among the top scorers, that is, the actual
probability of drawing the secret combination at each move. Using
always combinations of length equal to 4, we have proved that the
fact that Most Parts has an increased chance of finding the secret
combination among its top scorers counterbalances the effective
reduction of the set of consistent combinations brought by Entropy for
six colors; the difference among the probabilities for both method decreases with
the search space size (which we have proved for $\kappa=8$,
making Entropy beat Most Parts for that configuration.

Using that fact, we have proposed two new methods that effectively
combine Entropy and Most Parts by reducing search space size as the
former, and having a set of top scorers to choose at worst as bad as
Most Parts. These two methods, Plus and Plus2, behave statistically in
the same way for both sizes tested, are marginally (not significantly)
better than Most Parts for $\kappa=6, \ell=4$, and significantly
better than Most Parts and marginally better than Entropy for
$\kappa=8, \ell=4$, this method being itself only marginally better
than Most Parts for that size. Besides, results obtained by Plus/Plus2
are an improvement over the state of the art, published, for instance,
in \cite{DBLP:conf/cec/GuervosMC11} (which is comparable to that
obtained by Entropy and, thus, probably not statistically
significant). 

The two new methods presented do not present statistically significant
differences. This could be interpreted by stating that combining a scoring
strategy that reduces search space {\em and} gives top scores to the
hidden combination consistently, in general, is the most profitably
course of action. At the same time, doing to in different ways will
not lead to significant differences. However, although it is tempting
to generalize this to spaces of higher dimensions, heuristically it
cannot be done. We can, however, affirm that combining several scoring
strategies, specially Entropy and Most Parts, will yield better
results even if using samples of the whole consistent set. Besides,
both methods, but specially Plus2, is always (for the two sizes
tested, which is considered enough) as good as the best previous
method described, which makes it more robust.

This paper also introduces and tests a methodology for testing
different algorithms for solving Mastermind. While the first
approximation would be to empirically measure the average number of
moves, and the first would design methods that maximize the reduction
of search space every move, we have introduced here the measurement of
the probability of drawing the secret code as a third empirical test
to consider when designing a new method, since the success of a
solution depends, for the first moves, in that probability, to be
followed by the reduction of search space to a single element, as
considered so far.

In theory, these results and methodology could be extended to spaces with bigger sizes;
however, in practice, the increase in time complexity of the algorithm
bars it except from the simplest extensions. While solving  $\kappa=6,
\ell=4$ takes around 0.5s, $\kappa=8,\ell=4$, whose search space is
only 4 times as big, needs around 10 times as much time, around 5
seconds. One of the bottlenecks of the algorithm is the need to
compare all elements in the consistent set with each other, which is
approximately quadratic in size; this needs to be done for each
step, besides, the number of steps also increases in a complex way
with the increase in search space size. However, in time it will
become possible to test in a reasonable amount of time whether these
new methods are still better than the classical ones, and in which
proportion. It is impossible to know in advance what will be the
influence of finding the secret code among the top scorers; in fact,
it {\em increases} from  $\kappa=6$ to $\kappa=8$: the size of the
consistent set increases, but  the probability of finding
the secret code among them also increases (although the actual
probability of drawing that code decreases too); however, the actual size of
that set will also have an influence, with bigger sizes decreasing the
actual chance of playing the secret combination. How the three
quantities will change with the problem size (and, actually, with both
dimensions of the problem: number of colors $\kappa$ and combination
length $\ell$) is beyond the scope of this paper, and might not be easy
to compute analytically. 

It is more interesting, however, to use these results for methods that
use a sample of the set of consistent combinations, such as
evolutionary algorithms
\cite{DBLP:conf/evoW/GuervosCM11,Berghman20091880}. Results obtained
here can be used in two different ways: by taking into account the
size of the consistent set to set the sample size, which was set
heuristically \cite{nicso} or fixed \cite{Berghman20091880}, and also
by using a combination of Most Parts and Entropy scores to compute the
fitness of a particular combination. This will allow to find solutions
to Mastermind problems in bigger search spaces. However, the limiting
factor keeps on being the size of the consistent set that will be used
to score the combinations; since sample size increases with problem
size, eventually a non-feasible limit (in term of time or memory
usage) will be reached. However, using even a small size will extend
the range of problems with feasible solutions. Finding a way to score
solutions that is less-than-quadratic in time will also extend that
range, although another avenue, already explored years ago
\cite{jj-ppsn96}, would be to play non-consistent combinations in some
cases if enough time without finding consistent combinations passes.

This leaves as future work improving the scoring used in evolutionary
algorithms using these results, and checking how far these solutions
will go into the search space. Managing to find a solution to the
$\kappa=12,\ell=8$ in under one hour would be a good target, but it
will need a good adjustment of the evolutionary algorithm parameters,
as well as tweaking the implementation as far as it will allow
(because implementation matters
\cite{DBLP:conf/iwann/MereloRACML11}). Using either a distributed
computing environment such as the Evolvable Agents \cite{juanlu:SC08}
or SofEA \cite{sofea:naco} will probably be needed to in order to
shorten total time to solution. However, an parallel version is not
trivial, since things such as the consistent set might have to be centralized.

\section*{Acknowledgments}

 This work is supported by grants TIN2011-28627-C04-02, -01 and  P08-TIC-03903.

\bibliographystyle{elsarticle-num}
\bibliography{geneura,mastermind}

\begin{thebibliography}{10}
\expandafter\ifx\csname url\endcsname\relax
  \def\url#1{\texttt{#1}}\fi
\expandafter\ifx\csname urlprefix\endcsname\relax\def\urlprefix{URL }\fi
\expandafter\ifx\csname href\endcsname\relax
  \def\href#1#2{#2} \def\path#1{#1}\fi

\bibitem{mm:mathworld}
E.~W. Weisstein,
  \href{http://mathworld.wolfram.com/Mastermind.html}{Mastermind.}, From
  MathWorld--A Wolfram Web Resource.
\newline\urlprefix\url{http://mathworld.wolfram.com/Mastermind.html}

\bibitem{wiki:mm}
Wikipedia, \href{http://sl.ugr.es/001X}{Mastermind (board game) ---
  {W}ikipedia{,} {T}he {F}ree {E}ncyclopedia} (2009).
\newline\urlprefix\url{http://sl.ugr.es/001X}

\bibitem{francisstrategies:moo}
J.~Francis, {Strategies for playing MOO, or "Bulls and Cows"},
  \url{http://www.jfwaf.com/Bulls%20and%20Cows.pdf}.

\bibitem{Chen2007435}
S.-T. Chen, S.-S. Lin, L.-T. Huang, A two-phase optimization algorithm for
  mastermind, Computer Journal 50~(4) (2007) 435--443.

\bibitem{bank:mm}
R.~Focardi, F.~Luccio, Cracking bank pins by playing mastermind, in: P.~Boldi,
  L.~Gargano (Eds.), Fun with Algorithms, Vol. 6099 of Lecture Notes in
  Computer Science, Springer-Verlag, Berlin Heidelberg, 2010, pp. 202--213.

\bibitem{focardi2011guessing}
R.~Focardi, F.~Luccio, Guessing bank pins by winning a mastermind game, Theory
  of Computing Systems (2011) 1--20.

\bibitem{goodrich2009algorithmic}
M.~Goodrich, {On the algorithmic complexity of the Mastermind game with
  black-peg results}, Information Processing Letters 109~(13) (2009) 675--678.

\bibitem{gagneur2011selective}
J.~Gagneur, M.~Elze, A.~Tresch, Selective phenotyping, entropy reduction, and
  the mastermind game, BMC bioinformatics 12~(1) (2011) 406,
  \url{http://www.biomedcentral.com/1471-2105/12/406}.

\bibitem{mastermind05}
J.~J. Merelo-Guerv\'os, P.~Castillo, V.~Rivas, Finding a needle in a haystack
  using hints and evolutionary computation: the case of evolutionary
  {M}aster{M}ind, Applied Soft Computing 6~(2) (2006) 170--179,
  \url{http://www.sciencedirect.com/science/article/B6W86-4FH0D6P-1/2/40a99afa%
8e9c7734baae340abecc113a}; \url{http://dx.doi.org/10.1016/j.asoc.2004.09.003}.

\bibitem{stackoverflow}
{Several Authors}, How to solve the ''{Mastermind}'' guessing game?,
  \url{http://stackoverflow.com/questions/1185634/how-to-solve-the-mastermind-%
guessing-game}, question posed in StackOverflow and many answers.

\bibitem{o1991mastermind}
J.~O\'{}Geran, H.~Wynn, A.~Zhigljavsky, Mastermind as a test-bed for search
  algorithms, Chance 6 (1993) 31--37.

\bibitem{abs-cs-0512049}
J.~Stuckman, G.-Q. Zhang, \href{http://arxiv.org/abs/cs/0512049}{Mastermind is
  {NP}-complete}, INFOCOMP J. Comput. Sci 5 (2006) 25--28.
\newline\urlprefix\url{http://arxiv.org/abs/cs/0512049}

\bibitem{Kendall200813}
G.~Kendall, A.~Parkes, K.~Spoerer, A survey of {NP}-complete puzzles, ICGA
  Journal 31~(1) (2008) 13--34.

\bibitem{DBLP:journals/corr/abs-1111-6922}
G.~Viglietta, Hardness of mastermind, CoRR abs/1111.6922.

\bibitem{Kooi200513}
B.~Kooi, Yet another {M}astermind strategy, ICGA Journal 28~(1) (2005) 13--20.

\bibitem{jj-ppsn96}
J.~L. Bernier, C.-I. Herr\'aiz, J.-J. Merelo-Guerv\'os, S.~Olmeda, A.~Prieto,
  \href{http://www.springerlink.com/content/78j7430828t2867g}{{Solving {\em
  {M}aster{M}ind} using {GA}s and simulated annealing: a case of dynamic
  constraint optimization}}, in: Proceedings PPSN, Parallel Problem Solving
  from Nature IV, no. 1141 in Lecture Notes in Computer Science,
  Springer-Verlag, 1996, pp. 553--563.
\newblock \href {http://dx.doi.org/10.1007/3-540-61723-X\_1019}
  {\path{doi:10.1007/3-540-61723-X\_1019}}.
\newline\urlprefix\url{http://www.springerlink.com/content/78j7430828t2867g}

\bibitem{genmm99}
J.-J. Merelo-Guerv\'os, J.~Carpio, P.~Castillo, V.~M. Rivas, G.~Romero, Finding
  a needle in a haystack using hints and evolutionary computation: the case of
  {G}enetic {M}astermind, in: A.~S.~W. Scott~Brave (Ed.), Late breaking papers
  at the GECCO99, 1999, pp. 184--192.

\bibitem{Knuth}
D.~E. Knuth, The computer as {M}aster {M}ind, J. Recreational Mathematics 9~(1)
  (1976-77) 1--6.

\bibitem{Berghman20091880}
L.~Berghman, D.~Goossens, R.~Leus, Efficient solutions for {M}astermind using
  genetic algorithms, Computers and Operations Research 36~(6) (2009)
  1880--1885.

\bibitem{irving}
R.~W. Irving, Towards an optimum {M}astermind strategy, Journal of Recreational
  Mathematics 11~(2) (1978-79) 81--87.

\bibitem{Neuwirth}
E.~Neuwirth, Some strategies for {M}astermind, Zeitschrift fur Operations
  Research. Serie B 26~(8) (1982) B257--B278.

\bibitem{bestavros}
A.~Bestavros, A.~Belal,
  \href{http://citeseer.ist.psu.edu/bestavros86mastermind.html}{Mastermind, a
  game of diagnosis strategies}, Bulletin of the Faculty of Engineering,
  Alexandria University.
\newline\urlprefix\url{http://citeseer.ist.psu.edu/bestavros86mastermind.html}

\bibitem{mm:ppsn:2010}
C.~Cotta, J.~Merelo~Guerv\'os, A.~Mora~Garc\'ia, T.~Runarsson,
  \href{http://dx.doi.org/10.1007/978-3-642-15871-1\_43}{Entropy-driven
  evolutionary approaches to the {Mastermind} problem}, in: R.~Schaefer,
  C.~Cotta, J.~Kolodziej, G.~Rudolph (Eds.), Parallel Problem Solving from
  Nature PPSN XI, Vol. 6239 of Lecture Notes in Computer Science, Springer
  Berlin / Heidelberg, 2010, pp. 421--431.
\newline\urlprefix\url{http://dx.doi.org/10.1007/978-3-642-15871-1\_43}

\bibitem{nicso}
T.~P. Runarsson, J.~J. Merelo, Adapting heuristic {M}astermind strategies to
  evolutionary algorithms, in: NICSO'10 Proceedings, Studies in Computational
  Intelligence, Springer-Verlag, 2010, pp. 255--267, also available from ArXiV:
  \url{http://arxiv.org/abs/0912.2415v1}.

\bibitem{DBLP:conf/cec/GuervosMC11}
J.-J. Merelo-Guerv{\'o}s, A.-M. Mora, C.~Cotta, Optimizing worst-case scenario
  in evolutionary solutions to the {MasterMind} puzzle, in: IEEE Congress on
  Evolutionary Computation, IEEE, 2011, pp. 2669--2676.

\bibitem{DBLP:conf/evoW/GuervosCM11}
J.-J. Merelo-Guerv{\'o}s, C.~Cotta, A.~Mora, {Improving and Scaling
  Evolutionary Approaches to the MasterMind Problem}, in: C.~D. Chio,
  S.~Cagnoni, C.~Cotta, M.~Ebner, A.~Ek{\'a}rt, A.~Esparcia-Alc{\'a}zar,
  J.~J.~M. Guerv{\'o}s, F.~Neri, M.~Preuss, H.~Richter, J.~Togelius, G.~N.
  Yannakakis (Eds.), EvoApplications (1), Vol. 6624 of Lecture Notes in
  Computer Science, Springer, 2011, pp. 103--112.

\bibitem{DBLP:conf/iwann/MereloRACML11}
J.-J. Merelo-Guerv{\'o}s, G.~Romero, M.~Garc\'{\i}a-Arenas, P.~A. Castillo,
  A.-M. Mora, J.-L. Jim{\'e}nez-Laredo, Implementation matters: Programming
  best practices for evolutionary algorithms, in: J.~Cabestany, I.~Rojas, G.~J.
  Caparr{\'o}s (Eds.), IWANN (2), Vol. 6692 of Lecture Notes in Computer
  Science, Springer, 2011, pp. 333--340.

\bibitem{juanlu:SC08}
J.~L.~J. Laredo, P.~A. Castillo, A.~M. Mora, J.~J. Merelo, Evolvable agents, a
  fine grained approach for distributed evolutionary computing: walking towards
  the peer-to-peer computing frontiers, Soft Computing - A Fusion of
  Foundations, Methodologies and Applications 12~(12) (2008) 1145--1156.

\bibitem{sofea:naco}
J.-J. Merelo-Guerv\'os, A.~Mora, C.~Fernandes, A.~I. Esparcia, Designing and
  testing a pool-based evolutionary algorithm, submitted to Natural Computing
  (2012).

\end{thebibliography}

\end{document}